\documentclass{article}

\usepackage[preprint]{icml2026}

\usepackage[utf8]{inputenc} 
\usepackage[T1]{fontenc}    
\usepackage{hyperref}       
\usepackage{url}            
\usepackage{booktabs}       
\usepackage{amsfonts}       
\usepackage{nicefrac}       
\usepackage{microtype}      
\usepackage{bbm} 

\usepackage{amsmath}
\usepackage{amssymb}
\usepackage{mathtools}
 \usepackage[table]{xcolor} 
  \usepackage{graphicx} 

\usepackage{import}
\usepackage{amsmath}
\usepackage{algorithm}
\usepackage{graphicx}
\usepackage{subcaption}
\usepackage{booktabs}

\usepackage{algorithmic}

\usepackage{hyperref}
\usepackage{boxedminipage}
\usepackage{multirow}
\usepackage{pifont}
\usepackage{bm}
\usepackage{enumitem}
\usepackage{amsthm}      
\usepackage{thmtools} 
\usepackage{tikz}
\usetikzlibrary{positioning,shapes.geometric,arrows.meta}
\usetikzlibrary{shapes, arrows, positioning}
\usepackage[capitalize,noabbrev]{cleveref}

\theoremstyle{plain}

\theoremstyle{definition}

\theoremstyle{remark}

\newcommand{\E}{\mathbb{E}}

\usepackage[textsize=tiny]{todonotes}
\icmltitlerunning{DOME: Improving Signal-to-Noise in Stochastic Gradient Descent via Sharp-Direction Subspace Filtering}

\begin{document}
\twocolumn[
\icmltitle{DOME: Improving Signal-to-Noise in Stochastic Gradient Descent via Sharp-Direction Subspace Filtering}

\begin{icmlauthorlist}
\icmlauthor{Julien Nicolas}{ills,mila,mcgill,liris}
\icmlauthor{Mohamed Maouche}{inria}
\icmlauthor{Sonia Ben Mokhtar}{liris}
\icmlauthor{Mark Coates}{ills,mila,mcgill}
\end{icmlauthorlist}

\icmlaffiliation{ills}{International Laboratory on Learning Systems (ILLS), Montreal, Canada}
\icmlaffiliation{mila}{Mila – Quebec AI Institute, Montreal, Canada}
\icmlaffiliation{mcgill}{McGill University, Montreal, Canada}
\icmlaffiliation{inria}{Inria, INSA Lyon, CITI, UR3720, 69621 Villeurbanne, France}
\icmlaffiliation{liris}{CNRS, INSA Lyon, LIRIS, UMR 5205, 69621 Villeurbanne, France}

\icmlcorrespondingauthor{Julien Nicolas}{julien.nicolas@mail.mcgill.ca}

\icmlkeywords{Machine Learning, ICML}
\vskip 0.3in

]
\printAffiliationsAndNotice{}

\begin{abstract}
Stochastic gradients for deep neural networks exhibit strong correlations along the optimization trajectory, and are often aligned with a small set of Hessian eigenvectors associated with outlier eigenvalues. Recent work shows that projecting gradients away from this Hessian outlier subspace has little impact on optimization, despite capturing a large fraction of gradient variability. Since computing the Hessian is intractable in practice, we introduce a principled first-order characterization of the nuisance subspace based on the covariance of stochastic gradients, and propose an efficient method to estimate it online. We show that removing this subspace also has little impact on optimization, and yields practical benefits for applications sensitive to gradient signal-to-noise ratio such as gradient compression.
\end{abstract}

\section{Introduction}

Stochastic gradient descent methods remain the workhorse for training modern deep networks.
A line of recent work has documented that \emph{most} stochastic gradients are highly correlated across iterations and concentrate in a low-dimensional subspace that evolves slowly over training~\citep{gur2018gradient, azam2021recycling, li2022low}.
This observation has motivated practical mechanisms that reuse past gradient information to reduce costs. For example, by communicating low-rank updates in federated learning, or by leveraging temporal correlation to recycle or compress updates \citep{azam2021recycling, li2022low, vogels2019powersgd}. These results suggest that low-dimensional gradient structure is an optimization signal to preserve. 

At the same time, a different set of analyses paints a more nuanced picture. Empirically and theoretically, as batches get smaller compared to the dataset size and stochastic gradient descent moves away from full gradient descent, those gradients tend to align with a small set of stable directions associated with outlier eigenvalues of the Hessian \citep{papyan2019measurements, arous2023high, song2024does}. The number of those outlier directions seems comparable to the number of classes in classification problems and persists across architectures and datasets~\citep{gur2018gradient, papyan2019measurements, song2024does}.
Crucially, recent evidence shows that \emph{removing} the component of the (minibatch) gradient in this outlier subspace has little effect on convergence or final accuracy \citep{song2024does}.
This suggests that the most prominent low-dimensional directions in the gradient trajectory can correspond to persistent, high-variance structure that is weakly informative for descent and can be interpreted as a nuisance subspace rather than a signal subspace.

A practical obstacle is that identifying Hessian outliers during training is computationally intractable for modern models. Indeed, it requires computing high-dimensional second-order quantities over the whole training dataset.
Our starting point is therefore to ask: \emph{Can we define, estimate, and filter out a nuisance subspace using only first-order information?}
We propose a principled characterization based on the \emph{centered covariance} of stochastic gradients.
Intuitively, this covariance captures directions in which individual gradients fluctuate the most around their batch mean, and its leading eigenspace can be refined incrementally in the optimization procedure as new gradients are computed.
We introduce \textbf{DOME}, a method that maintains an estimate of this subspace projector using efficient randomized numerical linear algebra techniques, and filters gradients by projecting them onto the orthogonal complement of the nuisance subspace before applying downstream operations.
Filtering nuisance directions improves the signal-to-noise ratio of gradient-based updates and could be especially valuable in settings that are sensitive to gradient distortion.
For instance, differential privacy requires clipping and adding noise calibrated to gradient norms~\citep{abadi2016deep, dwork2006differential}, and gradient compression further perturbs updates during communication \citep{alistarh2017qsgd, vogels2019powersgd}.
By removing the component of the gradient that inflates norms without materially contributing to optimization, DOME can improve resilience to these distortions.
We show empirically that projecting away the covariance-defined nuisance component does not hurt and can even improve convergence speed and accuracy, while substantially improving performance in high-noise regimes and under aggressive compression.

\paragraph{Contributions.}
This work provides a first-order, online approach to identify and remove structured high-variance directions that dominate stochastic gradients without contributing proportionally to useful descent. Our main contributions are:
\begin{itemize}[leftmargin=*]
    \item \textbf{A first-order surrogate for sharp curvature directions.}
    We point out that the \emph{centered covariance} of stochastic gradients provides a principled first-order characterization of dominant harmful high-variance directions observed during training. This covariance is closely related to the Gauss--Newton component of the Hessian, whose leading eigenspace is known to align with stochastic gradients throughout training.

    \item \textbf{Online estimation via a streaming power method.}
    We propose to use an efficient, low-memory procedure that tracks the leading eigenspace of the centered gradient covariance from streaming minibatches, without forming or storing full $d\times d$ covariance matrices.

    \item \textbf{Gradient filtering to improve signal-to-noise.}
    We propose \textbf{DOME}, which filters stochastic gradients by projecting away the estimated nuisance subspace. We empirically show that this filtering does not harm learning in the standard setting despite removing most of the gradient norm. We use aggressive compression as an example of settings sensitive to gradient signal-to-noise ratio and show that DOME improves robustness in that case.
\end{itemize}

\paragraph{Paper organization.}  \Cref{sec:method} formalizes the problem, connects sharp directions to the Gauss--Newton/Fisher Hessian term, and motivates the centered gradient covariance as a first-order surrogate. \Cref{sec:algorithm} presents DOME and the streaming subspace update. \Cref{sec:experiments} evaluates the effect of filtering on learning dynamics and robustness under compression. \Cref{sec:related} discusses connections to low-rank gradient methods, curvature-aware optimization, and parameter-efficient fine-tuning, and we publish our code on an anonymous \href{https://anonymous.4open.science/r/ICML_subs_filter-5BFF/}{repository}.

\section{Method}\label{sec:method}

\subsection{Problem setting}

We consider supervised multi-class classification with dataset
\(\mathcal D=\{(x_i,y_i)\}_{i=1}^n\), model parameters
\(\bm\theta\in\mathbb R^d\), number of classes $C$ and cross-entropy loss
\(\ell(\bm\theta;x,y)\).
Training proceeds for \(T\) iterations using stochastic gradient descent or an
adaptive variant.

At iteration \(t\), we sample a minibatch \(\mathcal{B}_t=\{(x_{t,j},y_{t,j})\}_{j=1}^B \subset \mathcal{D}\) of size \(B\).

For each example \((x_{t,j},y_{t,j})\in\mathcal{B}_t\), we compute the per-example gradient
\(\nabla_{\bm\theta}\ell(\bm\theta_t; x_{t,j}, y_{t,j}) \in \mathbb{R}^d.
\)
The minibatch gradient is then
\begin{equation}
\bm g_t \;\triangleq\; \frac{1}{B}\sum_{j=1}^B \nabla_{\bm\theta}\ell(\bm\theta_t; x_{t,j}, y_{t,j})
\end{equation}
which provides an unbiased but noisy estimate of the population gradient
\(\nabla \mathcal L(\bm\theta_t)\), where
\(\mathcal L(\bm\theta)=\E[\ell(\bm\theta;x,y)]\).

Our objective is to filter the gradient update using only first-order information, in a way that preserves optimization performance (e.g. training loss dynamics and test accuracy) while improving
the signal-to-noise properties of gradients used by downstream mechanisms such as clipping, noise injection, compression or continual learning.
\subsection{Batch size--dependent alignment with sharp directions}
A robust empirical observation is that stochastic gradients remain strongly correlated throughout training and concentrate in a low-dimensional subspace ~\citep{gur2018gradient,vogels2019powersgd, hyeon2021fedpara,li2022low}.
However, the interpretation of this low-dimensional structure depends on the minibatch size.

When the minibatch size approaches the dataset size, stochastic gradients closely track the full-dataset gradient and optimization dynamics resemble full-batch gradient descent. The work of \citet{song2024does} showed that when the minibatch size decreases relative to the dataset size, the variance of stochastic gradients increases and optimization departs from this regime. In this small-batch setting, an alignment emerges between gradients and a small set of eigenvectors associated with outlier eigenvalues of the Hessian. These directions persist across training and often have dimension comparable to the number of classes in classification problems~\citep{papyan2019measurements}. They then showed that projecting gradients away from the full Hessian outlier subspace has little effect on convergence speed or final accuracy, despite removing a large fraction of gradient energy.

This indicates that the dominant low-dimensional structure induced by small minibatches can correspond to persistent, high-variance directions that are only weakly informative for long-term descent. These observations motivate treating the dominant sharp-direction subspace as a \emph{nuisance subspace} whose influence can be attenuated without harming
learning.
\subsection{Gauss--Newton decomposition and sharp directions}
We make the connection between sharp gradient directions and curvature explicit through the Gauss--Newton decomposition of the Hessian for cross-entropy classification.
Let $z_{\bm{\theta}}$ be a deep neural network parameterized by $\bm{\theta}$. Let $x$ denote a training sample, \(z_{\bm\theta}(x)\in\mathbb R^C\) the logits,
\(\bm p_{\bm\theta}(x)=\mathrm{softmax}(z_{\bm\theta}(x))\in\mathbb R^C\) the predicted class probabilities, and
\(\bm J_{\bm\theta}(x)=\nabla_{\bm\theta} z_{\bm\theta}(x) \in \mathbb{R}^{d \times C}\) the Jacobian of the logits with respect to the parameters.
Define the softmax covariance matrix~\citep{bishop2006pattern}
\begin{equation}
    \bm S_{\bm\theta}(x)
\;\triangleq\;
\mathrm{Diag}(\bm p_{\bm\theta}(x))
-
\bm p_{\bm\theta}(x)\bm p_{\bm\theta}(x)^\top
\;\in\;\mathbb R^{C\times C}.
\end{equation}

The population Hessian of the risk
\(\mathcal L(\bm\theta)=\E[\ell(\bm\theta;x,y)]\)
admits the decomposition~\citep{bishop2006pattern, botev2017practical}
\begin{align}
\bm H(\bm\theta)
&=
\underbrace{
\E_x\!\left[
\bm J_{\bm\theta}(x)^\top
\bm S_{\bm\theta}(x)
\bm J_{\bm\theta}(x)
\right]}_{\bm G(\bm\theta)\;\text{(Gauss--Newton / Fisher)}} \nonumber \\
&\quad+\;
\underbrace{
\E_{x,y}\!\left[
\sum_{c=1}^C
(p_{\bm\theta}(c\mid x)-\mathbbm{1}[y=c])\,
\nabla^2_{\bm\theta} z_{\bm\theta,c}(x)
\right]}_{\bm R(\bm\theta)\;\text{(remainder)}} .
\end{align}

The Gauss--Newton term \(\bm G(\bm\theta)\) depends on first-order derivatives of the model through the Jacobian, and on the second derivative of the loss with respect to the logits, which for softmax cross-entropy reduces to the covariance \(\bm S_{\bm\theta}(x)\).
Prior empirical loss-landscape analyses~\citep{sagun2016eigenvalues,sagun2017empirical,papyan2018spectrum,ghorbani2019investigation} indicate that the Hessian outliers observed in practice are closely tied to this Gauss--Newton/Fisher component, and that stochastic gradients tend to align overall with the Hessian leading eigenspace. Moreover, for a classification task with $C$ classes, there are on the order of $C^2$ Hessian outliers in the Gauss--Newton component, with $C$ outliers being associated with much larger eigenvalues~\citep{papyan2019measurements}.


\subsection{Centered gradient covariance and stochastic deviations}

We now turn to a first-order term that plays a central role in our approach:
the centered covariance of stochastic gradients.
Writing the stochastic per-sample gradient as
\begin{equation}
    \nabla_{\bm\theta}\ell(\bm\theta;x,y)
=
\nabla \mathcal L(\bm\theta_t) + \bm\xi_t,
\qquad
\E[\bm\xi_t]=\bm 0,
\end{equation}
with $\bm\xi_t$ being the gradient noise, we define the centered gradient covariance
\begin{align}
    \bm\Sigma_t
\;&\triangleq\;
\E\!\left[
(\nabla_{\bm\theta}\ell(\bm\theta;\cdot)-\nabla \mathcal L(\bm\theta_t))
(\nabla_{\bm\theta}\ell(\bm\theta;\cdot)-\nabla \mathcal L(\bm\theta_t))^\top
\right] \nonumber \\
&=
\E[\bm\xi_t\bm\xi_t^\top].
\end{align}
For cross-entropy classification, the per-sample gradient admits the form
\begin{equation}
    \nabla_{\bm\theta}\ell(\bm\theta;x,y)
=
\bm J_{\bm\theta}(x)^\top
\bigl(\bm p_{\bm\theta}(x)-\bm e_y\bigr),
\end{equation}
where \(\bm e_y\) denotes the one-hot encoding of the label~\citep{murphy2012machine}.
Conditioned on \(x\), the randomness in \(y\) induces variability in
\(\bm p_{\bm\theta}(x)-\bm e_y\), whose covariance is precisely
\(\bm S_{\bm\theta}(x)\).
The full (uncentered) covariance is then 
\begin{equation}
\label{cov_matrix_jacob}
    \bm\Sigma_t'
=
\E_x\!\left[
\bm J_{\bm\theta_t}(x)^\top
\bm S_{\bm\theta_t}(x)
\bm J_{\bm\theta_t}(x)
\right],
\end{equation}
which coincides with the Gauss--Newton component of the population
Hessian for softmax cross-entropy models
\citep{botev2017practical}.
\paragraph{Small-batch regime.}
When minibatches are small compared to the dataset size, stochastic fluctuations dominate the batchwise gradient~\citep{smith2018bayesian, bottou2018optimization}, so that
\(
\|\bm\xi_t\| \gg \|\nabla \mathcal L(\bm\theta_t)\|.
\)
In this regime, the uncentered second moment
\(\E[\nabla_{\bm\theta}\ell(\bm\theta;x,y) \nabla_{\bm\theta}\ell(\bm\theta;x,y)^\top]\) is well approximated by the centered covariance
\(\bm\Sigma_t\), and the contribution of
\(\nabla \mathcal L(\bm\theta_t)\nabla \mathcal L(\bm\theta_t)^\top\) is
negligible.
As a result, the dominant eigenspaces of \(\bm\Sigma_t\) and
\(\bm G(\bm\theta_t)\) coincide, which means that the deviations of the minibatch gradient around the true gradient are aligned with sharp directions of the Gauss--Newton component.

From our perspective, the role of \(\bm\Sigma_t\) is therefore not to approximate curvature, but to identify a low-dimensional subspace capturing persistent, high-variance \emph{deviations around the true gradient}, which could lead to oscillations around the ideal optimization trajectory. 

\paragraph{Connection to momentum as implicit low-pass filtering.}
\citet{song2024does} observe that momentum-based optimizers yield updates whose energy is less concentrated in the Hessian-outlier subspace and more spread in its bulk compared to non-momentum-based optimizers.
We remark that this is consistent with the view that outlier-aligned components can correspond to persistent, high-variance oscillations around the true (mean over the dataset) descent direction:
the exponential moving average (EMA) acts as a temporal low-pass filter, attenuating rapidly varying components.
However, EMA provides only an implicit and frequency-dependent filter: its effective cutoff depends on its weighting factor (e.g., $\beta_1$), which does not necessarily match the frequencies of nuisance oscillations. Moreover, it could potentially filter out relevant high frequency directions.
\subsection{Slow evolution of the nuisance subspace}
As opposed to prior work, we want to estimate the outlier subspace using first-order, stochastic quantities.
The effectiveness of tracking the dominant eigenspace of $\bm\Sigma_t$ from stochastic gradients relies on the fact that this eigenspace evolves slowly over training, so that successive estimates have substantial overlap. This property follows from the combined stability of the Jacobian and the smooth dependence of the softmax covariance on the model parameters. Recall that, for cross-entropy classification, the centered covariance of stochastic gradients can be written as Equation~\ref{cov_matrix_jacob}.

\paragraph{Stability of the Jacobian.}
In sufficiently overparameterized networks trained with small learning rates, it has been shown that parameter updates induce only small relative changes in the Jacobian $\bm J_{\bm\theta_t}(x)$ across iterations~\citep{jacot2018neural}.
This behavior is formalized by the Neural Tangent Kernel (NTK) theory, which shows that in the infinite-width limit of neural networks, the Jacobian remains constant along training. 
Empirical studies further indicate that a near stationarity persists well beyond the infinite-width limit under standard optimization settings~\citep{lee2019wide}. This suggests that, in this regime, rapid changes in $\bm\Sigma_t$ are unlikely to be driven by rapid Jacobian drift.

\paragraph{Smooth variation of the softmax covariance.}
The remaining source of temporal variation in $\bm\Sigma_t$ arises from the softmax covariance $\bm S_{\bm\theta_t}(x)$. Crucially, $\bm S_{\bm\theta}(x)$ depends on the parameters only through the logits $z_{\bm\theta}(x)$, and is a smooth matrix-valued function of these logits. Since the softmax map has bounded first and second derivatives, there exists a constant $L>0$ such that, for any two logit vectors $z,z'$,
\(
\|\bm S(z)-\bm S(z')\|
\;\le\;
L\,\|z-z'\|.
\)

Under standard training regimes with small learning rates, a single SGD update induces only a small change in the logits for any fixed input $x$~\citep{hardt2016train, sagun2017empirical}, implying that $\bm S_{\bm\theta_{t+1}}(x)$ is a small perturbation of $\bm S_{\bm\theta_t}(x)$ in operator norm.
As a result, by classical eigenvector perturbation results, in particular the Davis--Kahan theorem, the eigenspaces of $\bm S_{\bm\theta_t}(x)$ vary continuously under small operator norm perturbations, and successive eigenspaces have substantial overlap~\citep{davis1970rotation,stewart1990matrix}. The slow evolution argument is also supported by other studies of the Gauss--Newton component~\citep{botev2017practical}.

We note that perfect stability of the nuisance subspace is less critical than stability of the informative subspace for low-rank approaches~\citep{vogels2019powersgd}. Indeed, low-rank approaches aim to compute a basis in which the learning signal can be contained. If the true signal subspace drifts slowly during training but the measured basis is not effectively synchronized, then the learning process is interrupted~\citep{song2024does}. In contrast, if the nuisance subspace projector is out-of-sync with the true subspace projector, then the learning process becomes noisier but is not interrupted.

Overall, the nuisance subspace stability justifies estimating the nuisance subspace online from stochastic gradients. 

We now turn to our proposed streaming nuisance subspace estimation procedure. 

\section{Algorithm}\label{sec:algorithm}
We outline the complete DOME algorithm in Section~\ref{alg:DOMEclipping}. DOME maintains a hidden low-rank subspace projector that is updated online from newly computed gradients. This projector is then used to filter the minibatch gradient before it is passed to the downstream optimizer (and, when relevant, clipping, noise addition, or compression).

\paragraph{Overview.}
At each iteration \(t\), DOME performs two steps:
(i) update the nuisance subspace estimate from the current minibatch gradients;
(ii) form the minibatch gradient and filter it by projecting away its nuisance subspace component before applying the optimizer update.

\paragraph{(i) Online subspace estimation (streaming power method + QR).}
DOME tracks the dominant eigenspace of the \emph{centered} within-minibatch gradient covariance using only matrix--vector products, without forming any
\(d\times d\) covariance matrix, using the streaming randomized power method introduced in~\citep{yang2018history}.

Given per-example gradients in a mini-batch \(\{\bm g_{t,j}\}_{j=1}^B\), define the batch mean
\(
\bm\mu_t \;=\; \frac{1}{B}\sum_{j=1}^{B}\bm g_{t,j},
\)
and the centered gradient matrix
\begin{equation}
    \bm H_t
=
[\bm g_{t,1}-\bm\mu_t,\dots,\bm g_{t,B}-\bm\mu_t]
\;\in\;\mathbb R^{d\times B}.
\end{equation}
Let \(\bm U_{t-1}\in\mathbb R^{d\times k}\) denote the current orthonormal basis
for the nuisance subspace and \(\bm\Lambda_{t-1}\in\mathbb R^{k\times k}\) the covariance eigenvalues.
The covariance action on \(\bm U_{t-1}\) is computed as
\begin{equation}
    \bm Y_t'=\frac{1}{B}\bm H_t(\bm H_t^\top \bm U_{t-1}),
\end{equation}
which costs \(O(dBk)\) and avoids forming \(\bm H_t\bm H_t^\top\) explicitly which would require \(O(d^2)\) memory.

We combine this minibatch information with a running historical estimate through
a streaming covariance action update of the form
\begin{equation}
    \bm Y_t
=
\frac{t-1}{t}\bm U_{t-1}\bm\Lambda_{t-1}
+
\frac{1}{t}\cdot \frac{1}{B}\bm Y_t'.
\end{equation}
Finally, we orthonormalize the updated range using the Gram-Schmidt QR decomposition, i.e.
\(
\bm U_t,\bm R_t \leftarrow \mathrm{QR}(\bm Y_t),
\)
and extract the diagonal scaling \(\bm\Lambda_t\) from column norms of \(\bm Y_t\) (Algorithm~\ref{alg:UpdateNuisanceSubspace}).

\textbf{Note on computation and memory:}
The additional memory required to store the nuisance subspace, \(\bm U_t \in \mathbb{R}^{d \times k}\) (and its diagonal scaling \(\bm\Lambda_t\)), is reasonable in typical training regimes. In standard PyTorch training, backpropagation already requires \(\mathcal{O}(B d)\) memory to store intermediate activations for a mini-batch of size \(B\). In practice, \(B\) is commonly set to \(128\)–\(256\), or even larger, so that for the values of interest \(k = 10\)–\(100\), the projector storage cost is \(d k =  \mathcal{O}(B d)\) and does not constitute a prohibitive overhead. If the model dimension is prohibitive for the QR computation, nuisance directions may be estimated layer-wise. 

\paragraph{(ii) Gradient filtering.}
After updating \(\bm U_t\), we form the minibatch gradient
\(
\bm g_t=\frac{1}{B}\sum_{j=1}^B \bm g_{t,j}
\)
and filter it by projecting away the estimated nuisance subspace component:
\begin{equation}
    \tilde{\bm g}_t
=
(\bm I-\bm U_t\bm U_t^\top)\,\bm g_t.
\end{equation}
The filtered gradient \(\tilde{\bm g}_t\) is then passed unchanged to the base
optimizer (e.g., SGD) and to any downstream processing such as clipping,
noise addition, or compression.

\begin{algorithm}[htb]
\caption{DOME}
\label{alg:DOMEclipping}
\begin{algorithmic}[1]
\STATE \textbf{Input:} Initial parameters \(\bm\theta_0\), learning rate \(\eta\), minibatch size \(B\), number of iterations \(T\),
Adam hyperparameters \(\beta_1,\beta_2\), numerical stabilizer \(\gamma'\), subspace rank \(k\).
\STATE \textbf{Initialize:} \(\tilde{\bm m}_0=\bm 0,\ \tilde{\bm v}_0=\bm 0\).
Initialize subspace basis \(\bm U_0\in\mathbb R^{d\times k}\) (QR on random Gaussian matrix),
and eigenvalues \(\bm\Lambda_0=\bm I_k\).
\FOR{$t=1,\dots,T$}
    \STATE Sample a minibatch \(\mathcal B_t\) of size \(B\).
    \STATE \(\bm g_{t,j}\leftarrow \nabla \ell(\bm\theta_{t-1};x_j)\) for all \(x_j\in\mathcal B_t\).

    \vspace{0.3em}
    \STATE \(\bm U_t,\bm\Lambda_t \leftarrow
    \texttt{UpdateSub}(\bm U_{t-1},\bm\Lambda_{t-1},\{\bm g_{t,j}\}_{j=1}^B)\)

    \vspace{0.3em}
    \STATE \textbf{// Form batch gradient and filter nuisance directions}
    \STATE \(\bm g_t \leftarrow \frac{1}{B}\sum_{j=1}^B \bm g_{t,j}\).
    \STATE \(\tilde{\bm g}_t \leftarrow (\bm I-\bm U_t\bm U_t^\top)\bm g_t\).
    \vspace{0.3em}
    \STATE \textbf{// Optimizer update}
    \STATE \(\tilde{\bm m}_t \leftarrow \beta_1 \tilde{\bm m}_{t-1} + (1-\beta_1)\tilde{\bm g}_t\),
    \(\ \hat{\bm m}_t \leftarrow \tilde{\bm m}_t/(1-\beta_1^t)\).
    \STATE \(\bm\theta_t \leftarrow \bm\theta_{t-1} - \eta \cdot \hat{\bm m}_t\).
\ENDFOR
\end{algorithmic}
\end{algorithm}

\begin{algorithm}[htb]
\caption{UpdateSub}
\label{alg:UpdateNuisanceSubspace}
\begin{algorithmic}[1]
\STATE \textbf{Input:} Orthonormal nuisance basis
\(\bm U_{t-1}\in\mathbb R^{d\times k}\),
eigenvalue proxy \(\bm\Lambda_{t-1}\in\mathbb R^{k\times k}\),
minibatch gradients
\(\{\bm g_{t,j}\}_{j=1}^B \subset \mathbb R^d\),
iteration \(t\).
\STATE \textbf{Parameter:} \(k\).

\vspace{0.2em}
\STATE \textbf{// Batch-wise mean gradient}
\STATE \(\bm\mu_t \leftarrow \frac{1}{B}\sum_{j=1}^{B}\bm g_{t,j}\)
\hfill (batch mean)

\vspace{0.2em}
\STATE \textbf{// Concatenated centered gradients}
\STATE Define \(\bm H_t \in \mathbb R^{d\times B}\) with columns
\(\bm h_{t,j} \leftarrow \bm g_{t,j}-\bm\mu_t\).

\vspace{0.2em}
\STATE \textbf{// Low-memory covariance mult.}
\STATE \(\bm V \leftarrow \bm H_t^\top \bm U_{t-1}\)
\hfill (\(\bm V\in\mathbb R^{B\times k}\))
\STATE \(\bm W \leftarrow \frac{1}{B}\bm H_t \bm V\)
\hfill (\(\bm W \in\mathbb R^{d\times k}\))

\vspace{0.2em}
\IF{$t=1$}
    \STATE \(\bm Y \leftarrow \bm W\)
    \hfill (initialize range)
\ELSE
    \STATE \(\bm Y \leftarrow \frac{t-1}{t}\bm U_{t-1}\bm\Lambda_{t-1} + \frac{1}{t} \bm W\)
    \hfill (streaming  update)
\ENDIF

\STATE \(\bm U_t,\bm R \leftarrow \texttt{QR}(\bm Y)\)
\hfill (orthonormalize)

\STATE \(\bm\lambda_t \leftarrow
\bigl(\|\bm Y_{:,1}\|_2,\;\|\bm Y_{:,2}\|_2,\;\dots,\;\|\bm Y_{:,k}\|_2\bigr)\)
\STATE \(\bm\Lambda_t \leftarrow \mathrm{diag}(\bm\lambda_t)\)

\STATE \textbf{return} \(\bm U_t,\bm\Lambda_t\)
\end{algorithmic}
\end{algorithm}

\section{Evaluation}\label{sec:experiments}
We empirically evaluate DOME on image classification benchmarks to assess (i) whether filtering the dominant eigenspace of the gradient covariance affects learning dynamics, and (ii) whether it improves robustness in settings where gradient signal-to-noise ratio (SNR) is critical. 

\subsection{Experimental Setup}
\paragraph{Datasets.}
We consider \textbf{MNIST} (70,000 images, $10$-class), \textbf{CIFAR-10} (60,000 images, $10$-class), and \textbf{TinyImageNet} (120,000 images, $200$-class).
MNIST consists of grayscale $28\times 28$ handwritten digits; CIFAR-10 contains $32\times 32$ RGB natural images; TinyImageNet contains $64\times 64$ RGB images. We use the standard train/test splits.

\paragraph{Model architecture.}
For CIFAR-10 and MNIST, we use a lightweight \textbf{ResNet-8}~\citep{song2024does} with GroupNorm ($7.8\times 10^4$ parameters).
For TinyImageNet we use a \textbf{ResNet-18} backbone with a 200-way classifier head ($ 1.13\times 10^7$ parameters). ResNet-8 enables fast ablations on CIFAR-10/MNIST, while ResNet-18 is a standard, competitive baseline for TinyImageNet~\citep{park2021influence, liu2022automix, amangeldi2025cnn}.

\paragraph{Training protocol.}
For each configuration, we repeat the training process with 5 different random seeds and report mean metrics with bootstrap confidence intervals.
For DOME, the nuisance subspace rank is set to $k=C^2$ (e.g., $k=100$ for $C=10$) for Figures \ref{fig:training_cifar10},\ref{app:compression_mnist},\ref{fig:covariance_spectrum}--\ref{app:cifar10_acc_vs_batchsize} and \ref{app:cifar10_training_multibs}. Due to the larger number of classes in TinyImageNet and to memory limitations, we set $k=C=200$ for Figure~\ref{fig:training_tinyimagenet}.
We use Opacus~\citep{yousefpour2021opacus} to enable efficient per-sample gradient computation required for the within-minibatch centered covariance updates.
Learning-dynamics figures on CIFAR-10 and TinyImageNet (Figs.~\ref{fig:training_cifar10}--\ref{fig:training_tinyimagenet}) use SGD with learning rate $0.1$ and momentum coefficient $\beta=0.9$.
Compression experiments (Figs.~\ref{fig:compression}--\ref{fig:compression_k_ablation}) use Adam with learning rate $10^{-3}$ and momentum coefficients $\beta_1=0.9$ and $\beta_2=0.999$.
We choose Adam in this setting because gradient compression alters the scale of the updates in a compression-rate--dependent manner, while Adam is approximately scale invariant due to its second-moment normalization.
Using Adam therefore allows us to avoid scale effects induced by compression and to isolate the impact of DOME on optimization performance.

\subsection{Experiments}
We first consider four main experiments.

\paragraph{(1) Learning dynamics with and without filtering.} \label{learndynamics}
We first study the effect of the dominant subspace filtering on standard training dynamics, without any additional gradient distortion, on image classification tasks using CIFAR-10 and TinyImageNet. We use a ResNet-8 with batch size 16 on CIFAR-10 and a ResNet-18 with batch size 64 on TinyImageNet. We further vary the batch sizes in Appendix~\ref{app:cifar10_acc_vs_batchsize}--\ref{app:cifar10_training_multibs} and we additionally validate our findings on a text classification task (DBPedia-Classes, L2 level with 70 classes) in Appendix~\ref{app:text_classification}.

\begin{figure*}[htb]
  \centering
  \begin{subfigure}[t]{0.33\linewidth}
    \centering
    \includegraphics[width=\linewidth]{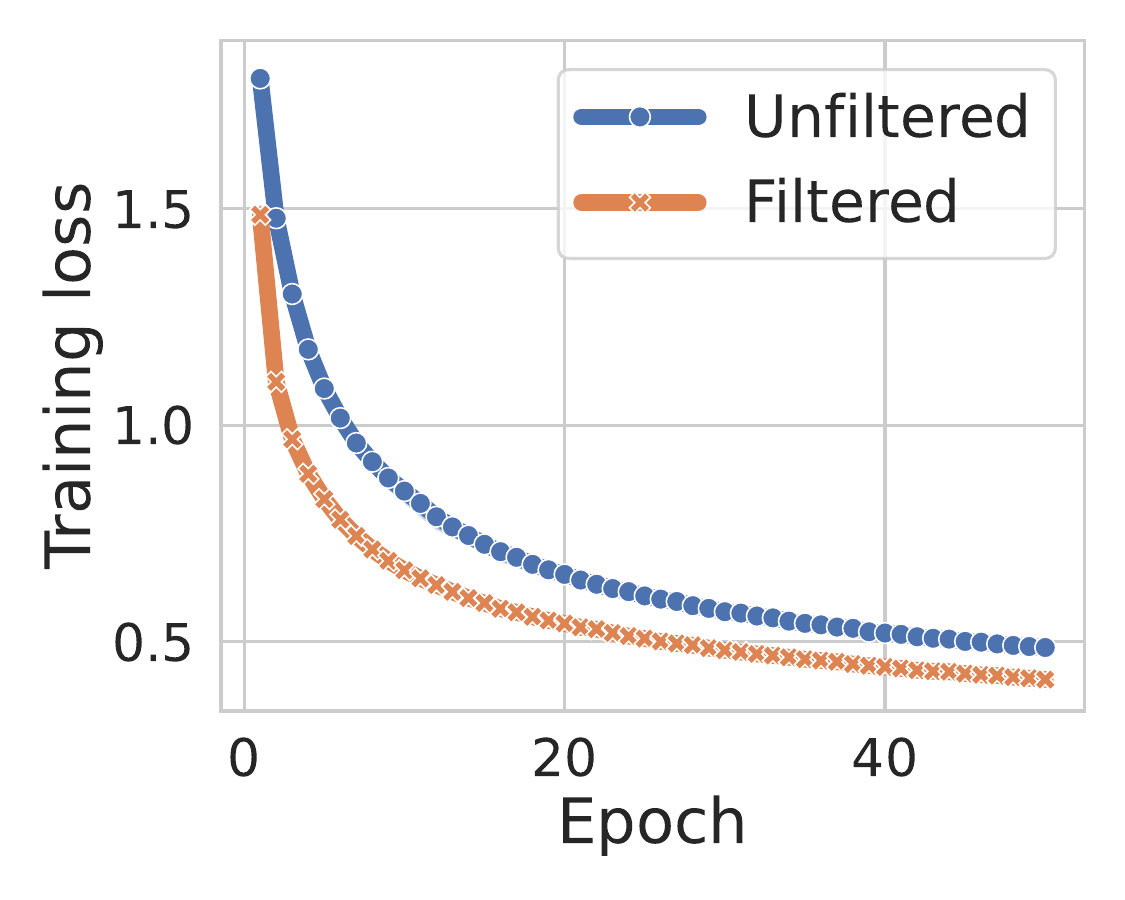}
    \label{fig:training_cifar10_loss}
  \end{subfigure}\hfill
  \begin{subfigure}[t]{0.33\linewidth}
    \centering
    \includegraphics[width=\linewidth]{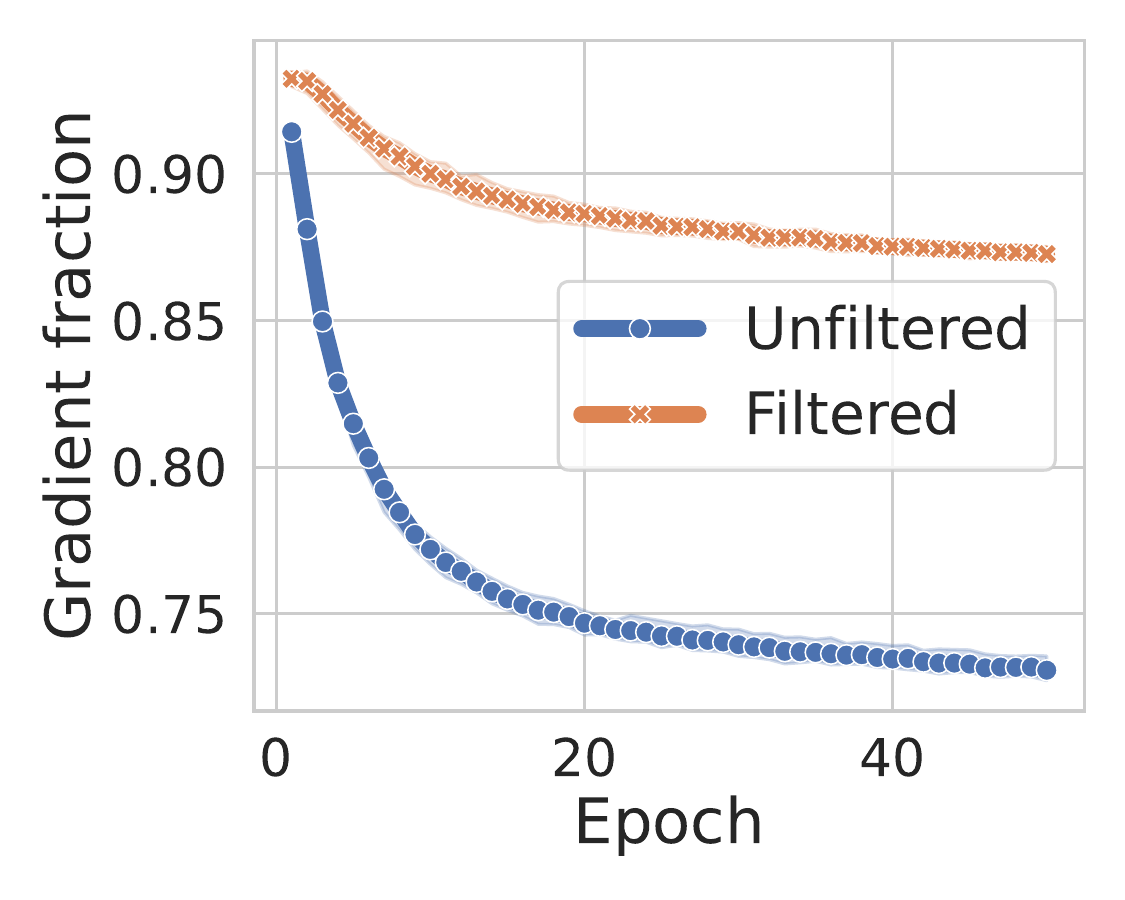}
    \label{fig:training_cifar10_fraction}
  \end{subfigure}\hfill
  \begin{subfigure}[t]{0.33\linewidth}
    \centering
    \includegraphics[width=\linewidth]{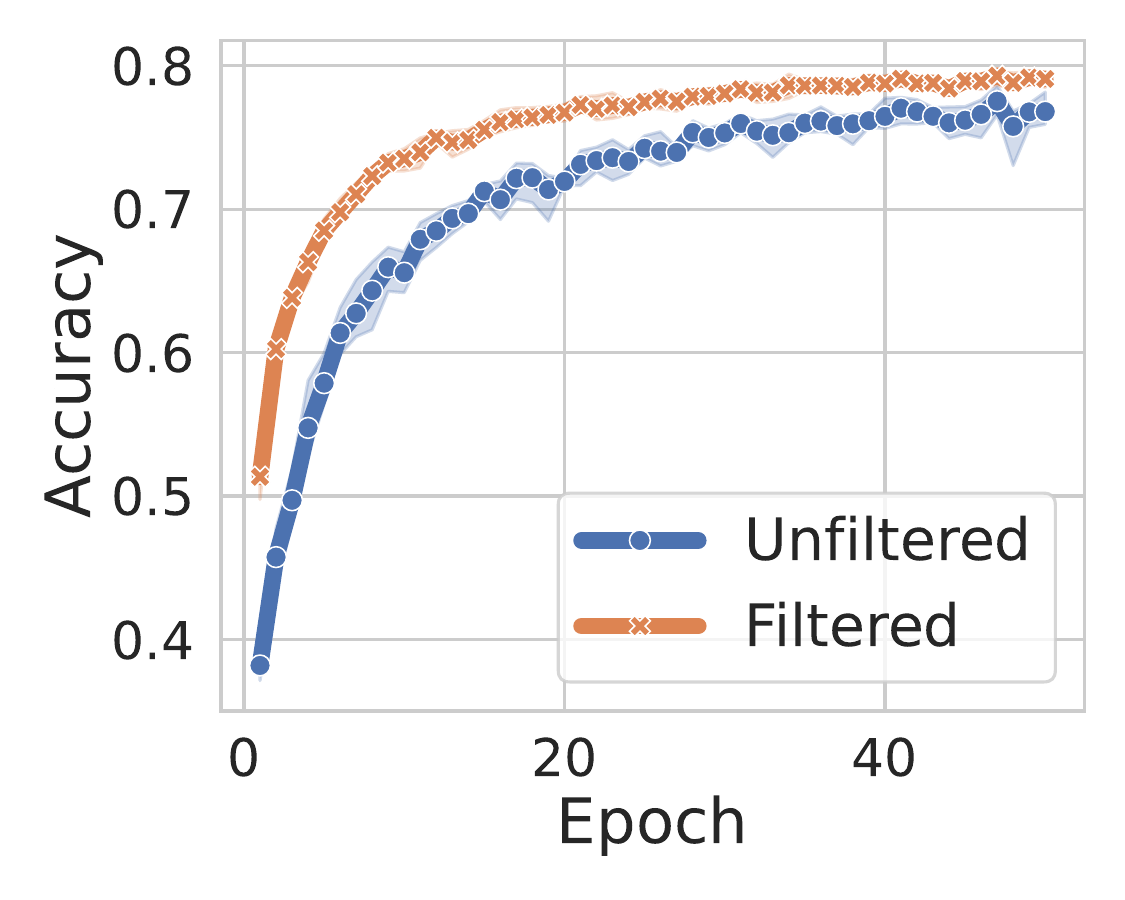}
    \label{fig:training_cifar10_accuracy}
  \end{subfigure}
  \vspace{-4mm}
  \caption{
  \textbf{Impact of filtering on training dynamics on CIFAR-10} when training a ResNet-8 with SGD, $lr=0.1$ and a batch size of 16 for 50 epochs.
  \textbf{Left}: Training loss as a function of epochs for the unfiltered optimizer and its filtered counterpart.
  \textbf{Center}: Average fraction of gradient norms lying in the dominant subspace (before applying filtering for the filtered version).
  \textbf{Right}: Top-1 accuracy as a function of epochs.
  Shaded areas indicate $99\%$ bootstrap confidence intervals over 5 random seeds.
  }
  \label{fig:training_cifar10}
    \vspace{-2mm}
\end{figure*}

\begin{figure*}[htb]
  \centering
  \begin{subfigure}[t]{0.33\linewidth}
    \centering
    \includegraphics[width=\linewidth]{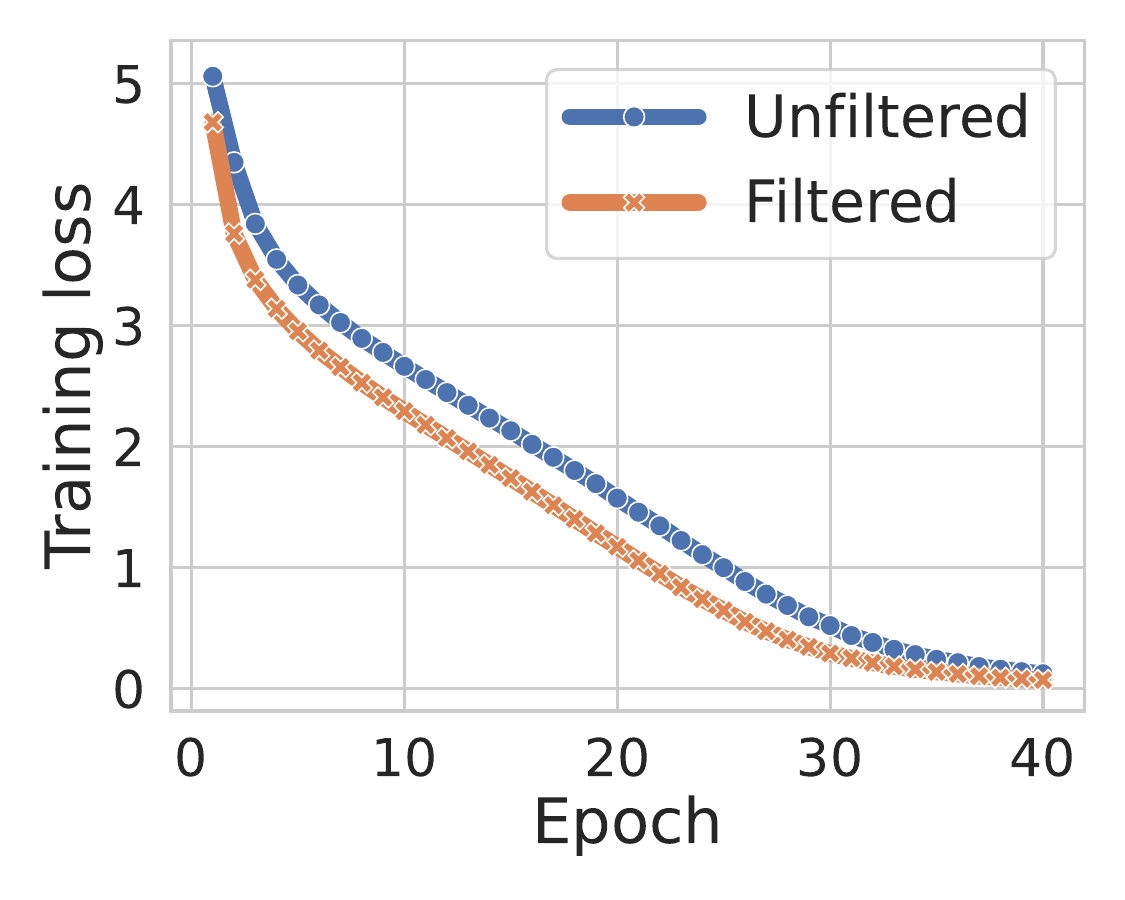}
    \label{fig:training_tinyimagenet_loss}
  \end{subfigure}\hfill
  \begin{subfigure}[t]{0.33\linewidth}
    \centering
    \includegraphics[width=\linewidth]{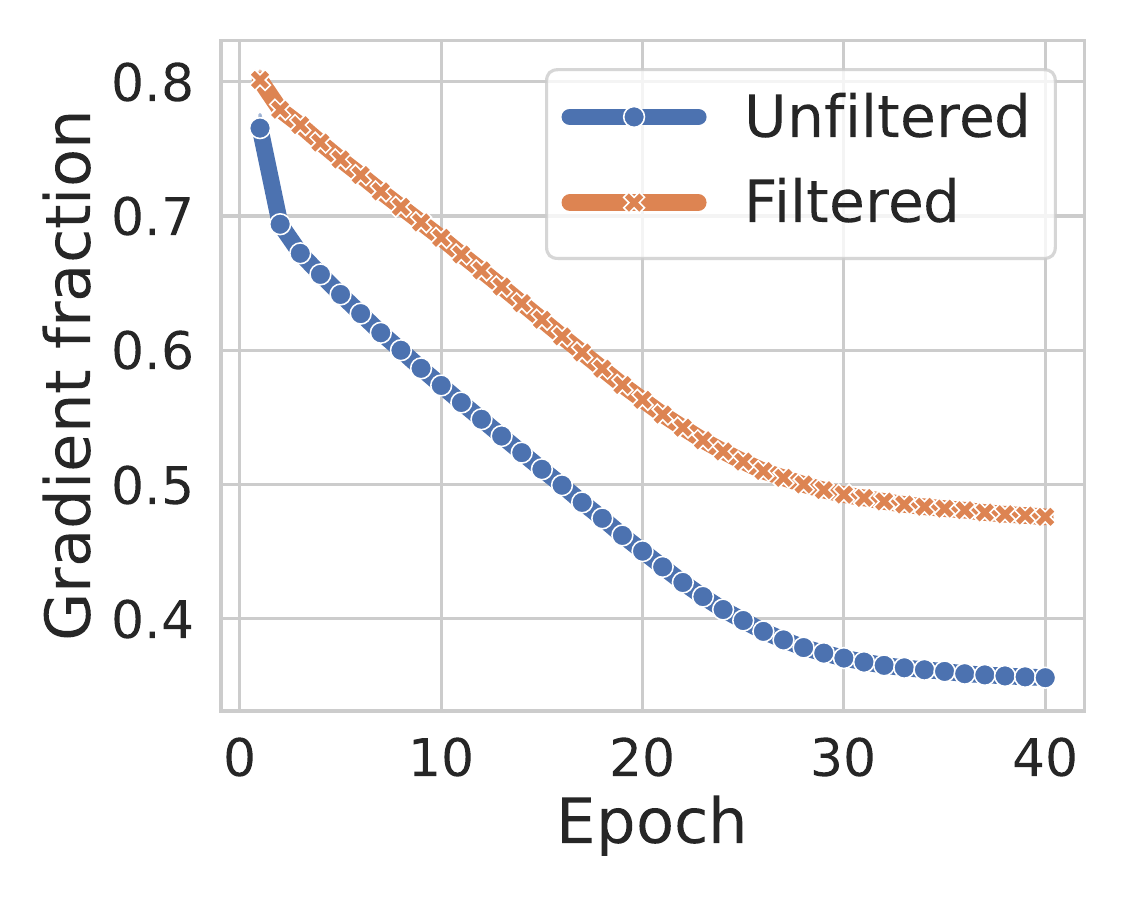}
    \label{fig:training_tinyimagenet_fraction}
  \end{subfigure}\hfill
  \begin{subfigure}[t]{0.33\linewidth}
    \centering
    \includegraphics[width=\linewidth]{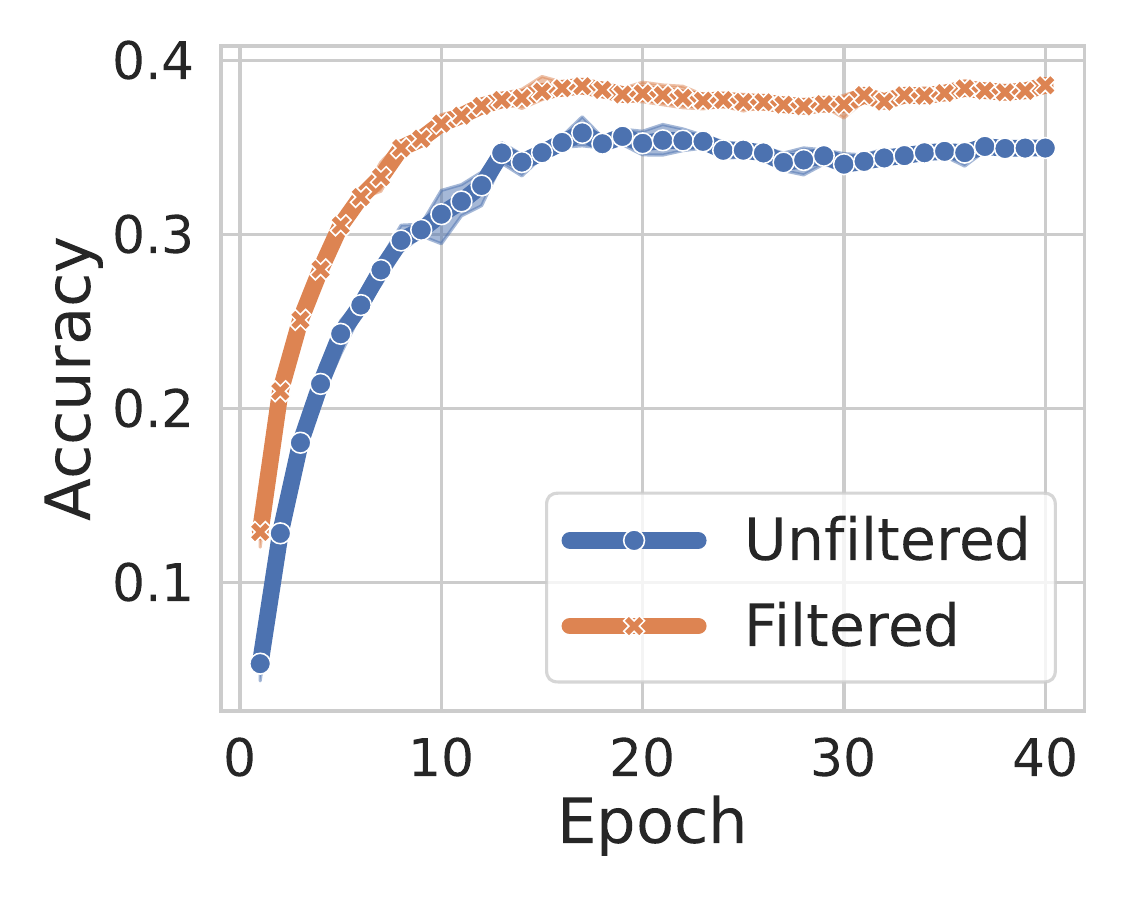}
    \label{fig:training_tinyimagenet_accuracy}
  \end{subfigure}
  \vspace{-6mm}
  \caption{
  \textbf{Impact of filtering on training dynamics on TinyImageNet} when training a ResNet-18 with SGD, $lr=0.1$, and a batch size of 64 for 50 epochs.
  \textbf{Left}: Training loss as a function of epochs for the unfiltered optimizer and its filtered counterpart.
  \textbf{Center}: Average fraction of gradient norms lying in the dominant subspace (before applying filtering for the filtered version).
  \textbf{Right}: Top-1 accuracy as a function of epochs.
  Shaded areas indicate $99\%$ bootstrap confidence intervals over 5 random seeds.
  }
  \label{fig:training_tinyimagenet}
    \vspace{-2mm}
\end{figure*}

To characterize the directions removed by DOME, Figures~\ref{fig:training_cifar10} and
\ref{fig:training_tinyimagenet} (middle) report the fraction of gradient norm lying in the
dominant covariance subspace, defined as the minibatch average of the per-sample gradient
norm captured in the nuisance subspace:
\[
\frac{1}{|\mathcal{B}_t|}\sum_{i \in \mathcal{B}_t}
\frac{\|\bm{U}_t \bm{U}_t^\top \bm g_{t,i}\|_2}{\| \bm g_{t,i}\|_2},
\]
where $\bm g_{t,i}$ denotes the per-sample gradient at step $t$ and $\bm{U}_t$ is the orthogonal
projector onto the dominant covariance subspace estimated at that step.

We also report the training loss trajectories (left) and the accuracies (right) across epochs in Figures~\ref{fig:training_cifar10} and~\ref{fig:training_tinyimagenet}. Along both unfiltered SGD trajectories, the evolution of the fraction of the gradient norm in the subspace seems to follow the evolution of the training loss, with Spearman correlations of $\rho_{1}=0.98$ and $\rho_{2}=0.96$, respectively. In other words, the norm of the gradient component in the nuisance subspace becomes smaller compared to the norm of the gradient itself as training progresses, suggesting that gradients get more and more aligned and that variance is reduced. 

We then observe that explicitly removing the component of the gradient in the dominant subspace at each step does not harm convergence speed and leads in fact to slightly smaller training loss and higher accuracy at a given step. This further indicates that the dominant subspace captures essentially \emph{noise}. However, the gradient fraction pre-filtering is not an indicator of the training progression itself, as it can be higher for the filtered runs (before applying filtering) than for the unfiltered runs, although the training loss can be smaller and accuracy higher.


\paragraph{(2) Gradient compression as a low-SNR application.}
We next evaluate DOME in a setting where gradient signal-to-noise ratio is directly critical:
\emph{gradient compression}.
Gradient compression introduces approximation error through lossy representations of the update. Its effect depends critically on how gradient energy is distributed across directions: when a large fraction of the gradient norm lies in high-variance but weakly informative components, these directions can dominate the compressed representation and lead to substantial distortion after reconstruction.

We consider a standard compression scheme based on \emph{random Gaussian projections} \citep{johnson1984extensions,stich2018sparsified, chen2022fundamental}.
At each optimization step \(t\), let \(\bm g_t \in \mathbb{R}^d\) denote the stochastic batchwise gradient. At each iteration, we draw independently a random projection matrix \(\bm R_t \in \mathbb{R}^{m \times d},
(\bm R_t)_{ij} \sim \mathcal{N}(0, 1/m).\)
The \emph{compressed-and-recovered gradient} is then defined as \(\bm g_t^{c}
\;=\;
\bm R_t^\top \bm R_t \bm g_t,\)
which corresponds to projecting \(\bm g_t\) onto a random \(m\)-dimensional subspace and reconstructing it back in \(\mathbb{R}^d\), with compression rate \(d/m\). 


In our experiments, this compression is applied at every iteration before the optimizer update for a total of 50 epochs with batch size 128. We then filter the gradient before compression, enabling a controlled comparison with Adam.

\begin{figure}[htb]
\centering
\includegraphics[width=0.95\linewidth]{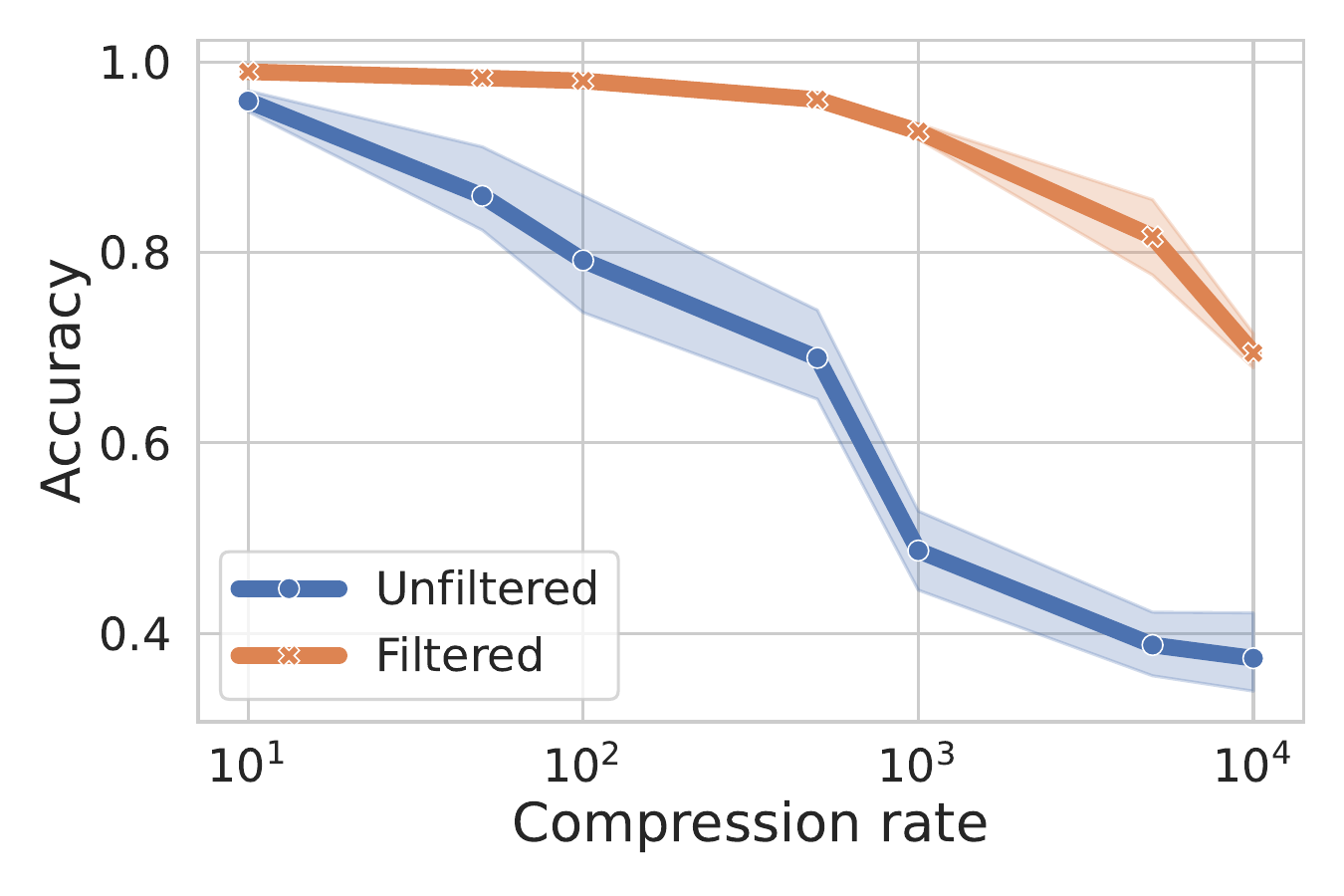}
\label{fig:compression_mnist}
    \vspace{-2mm}
  \caption{\textbf{Compression tolerance.} Test accuracy versus compression rate for Adam and DOME-filtered Adam using random Gaussian projection sketches, on MNIST with a ResNet-8. Training proceeds for 50 epochs with batch size 128. Curves report means over 5 runs; shaded regions indicate 95\%
  bootstrap confidence intervals.}
  \label{fig:compression}
    \vspace{-4mm}
\end{figure}

Figure~\ref{fig:compression} and Appendix~\ref{app:compression} report test accuracy as a function of the compression rate on MNIST and CIFAR-10. While the filtered and unfiltered methods perform similarly at low compression levels, standard Adam degrades rapidly as compression becomes more aggressive, whereas DOME maintains substantially higher accuracy. These results support our central claim: a large fraction of the gradient norm is concentrated in a low-dimensional nuisance subspace that is weakly informative for optimization but that can be amplified by downstream applications, e.g. compression. 

\begin{figure}[htb]
  \centering
  \includegraphics[width=0.95\linewidth]{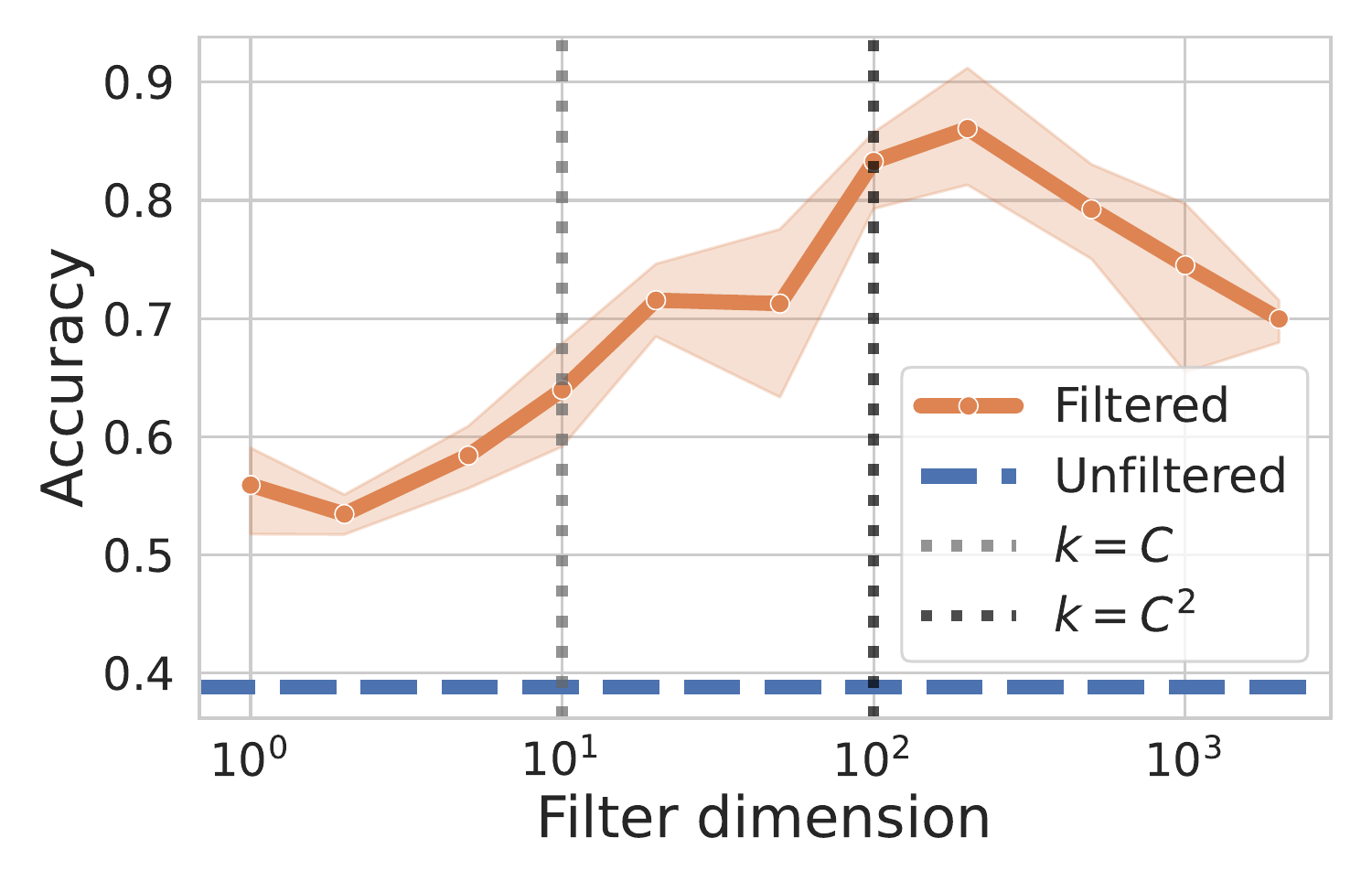}
  \vspace{-3mm}
  \caption{\textbf{Impact of the nuisance subspace dimension.}
  Test accuracy for Adam and DOME-filtered Adam under a fixed compression rate \(d/m=10^3\) for varying subspace rank $k$, on MNIST with a ResNet-8. Curves report means over 5 runs; shaded regions indicate 95\% bootstrap confidence intervals.}
  \label{fig:compression_k_ablation}
  \vspace{-3mm}
\end{figure}

\paragraph{(3) Impact of the nuisance subspace dimension.}
We finally study how the nuisance subspace dimension \(k\) impacts the benefits of filtering under aggressive compression. We fix a large compression rate \(d/m = 10^3\) and vary the rank of the filtered subspace \(
k \in \{1,2,5,10,20,50,100,200,500,1000,2000\}.
\).
All other hyperparameters are kept identical to experiment (2). Figure~\ref{fig:compression_k_ablation} reports the resulting test accuracy.

We observe that filtering yields a measurable improvement even for very small ranks (e.g., \(k\leq 10\)), suggesting that only a handful of dominant high-variance directions are sufficient to substantially mitigate compression distortion.
Performance improves as \(k\) increases up to \(k\approx C^2\), but overshooting this scale yields diminishing returns, with a degradation in performance for \(k \gg C^2\).
This behavior is consistent with the empirical picture from curvature analyses~\citep{papyan2018spectrum}: the dominant nuisance structure is concentrated in a subspace whose effective dimension is on the order of the outlier set (roughly \(C^2\) for cross-entropy classification with $C$ dominant outliers), while the remaining directions behave more like a high-dimensional bulk that needs to be preserved under compression. We note that \citet{halko2011finding} recommend using $k'=2k$ as the sketching dimension for the randomized power method if capturing the span of the top-$k$ eigenvectors is critical.

\paragraph{(4) Spectrum of the centered gradient covariance.}
Finally, we analyze the spectral structure of the centered gradient covariance
estimated by DOME.

\begin{figure}[htb]
    \centering
    \includegraphics[width=0.94\linewidth]{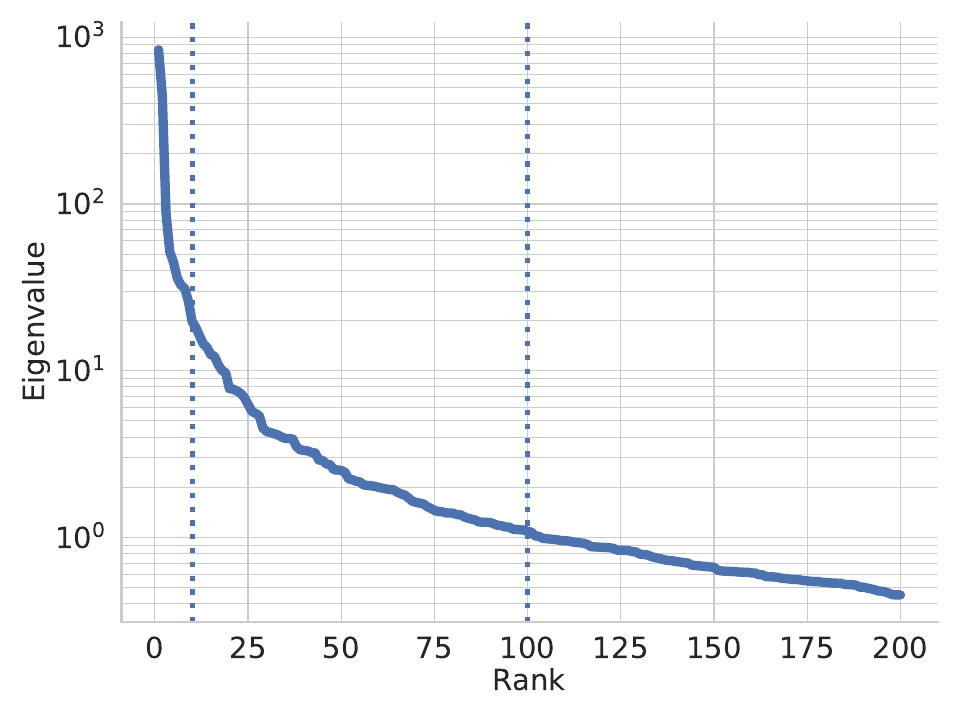}
    \label{fig:cifar10_spectrum}
    \vspace{-2mm}
  \caption{\textbf{Centered covariance spectrum.} Eigenvalue spectrum of the centered gradient covariance after training for 10 epochs on CIFAR-10 with batch size 128. The first dotted line is at rank $C=10$ and the
  second at rank $C^2=100$.}
  \label{fig:covariance_spectrum}
    \vspace{-3mm}
\end{figure}

Figure~\ref{fig:covariance_spectrum} shows the eigenvalue spectrum after training for 10 epochs on CIFAR-10. The spectrum for MNIST is deferred to Appendix~\ref{app:cov_spectrum}. The measured spectrum exhibits a similar structure that has been reported for the Hessian and Gauss--Newton matrices in empirical loss landscape studies~\citep{papyan2018spectrum,papyan2019measurements}: (i) a small number of very large eigenvalues whose count closely matches the number of classes
($C=10$), (ii) a broader set of additional outliers on the order of $C^2$, and (iii) a relatively flat bulk of small eigenvalues.
The fact that this structure emerges from a purely first-order statistic supports using the leading eigenspace of the centered covariance as a practical surrogate for identifying nuisance directions during training.

\section{Related Work}\label{sec:related}
\paragraph{Gradient quantization and compression.}
Quantization methods such as QSGD~\citep{alistarh2017qsgd,zhou2017incremental} and TernGrad~\citep{wen2017terngrad} reduce the precision of gradient coordinates, while sparsification techniques transmit only the largest components, often with error compensation to preserve convergence \citep{lin2018deep,karimireddy2019error,liurethinking,jiang2022model}.

\paragraph{Exploiting Gradient Structure for Efficiency.}
Low-rank structure in parameters or individual gradient updates (not across iterations) has been primarily leveraged to reduce communication. 
Low-rank approaches such as PowerSGD~\citep{vogels2019powersgd} or FedPara~\citep{hyeon2021fedpara} explicitly approximate individual gradients or updates using rank-constrained representations. 
Although developed in a federated context, these methods fundamentally rely on the assumption that gradient updates lie in a low-dimensional subspace that captures the essential optimization signal. Low-rank adaptation methods (e.g. LoRA~\citep{hu2022lora}, GLoRA~\citep{chavan2023one}) impose a low-rank parameterization of the update during fine-tuning by restricting weight changes to a rank-$r$ factorization. In contrast, DOME tries to alleviate the spurious temporal gradient correlation and is not a replacement for these methods, but as an orthogonal filtering step that targets a specific failure mode.

\paragraph{Temporal redundancy in gradient dynamics.}
A recent body of empirical work has documented that stochastic gradients in deep networks are strongly temporally correlated and often concentrate in a slowly varying low-dimensional subspace along training \citep{gur2018gradient,li2022low}.
These observations have motivated a range of methods in federated settings that recycle information across iterations by communicating differences between the current gradient and a reference from previous rounds \citep{azam2021recycling}. In contrast to DOME, existing approaches implicitly assume that the dominant subspace is informative and should be emphasized during training while we build on the view that this subspace is not beneficial for optimization.

\paragraph{Hessian outlier analyses.}
Analyses of the loss landscape report that the Hessian exhibits a small number of outlier eigenvalues separated from a flat bulk \citep{sagun2016eigenvalues,sagun2017empirical,papyan2018spectrum,ghorbani2019investigation}. Recent analysis shows that projecting away gradients from the top gradient subspace does not hurt optimization~\citep{song2024does}.

\paragraph{Curvature and sharpness aware optimization.}
A broad line of work leverages curvature information to improve optimization in deep networks. Natural gradient methods account for the geometry induced by the loss via the Fisher information \citep{amari1998natural}, with scalable approximations such as K-FAC exploiting layerwise structure in Gauss–Newton or Fisher matrices~\citep{martens2015optimizing,grosse2016kronecker}. Other second-order and quasi-second-order approaches mitigate sharp curvature effects through preconditioning or diagonal Hessian estimates~\citep{botev2017practical,yang2023adahessian}, while sharpness-aware methods explicitly modify the objective to dampen sharp directions~\citep{keskar2017large,foret2021sharpnessaware}. In contrast, DOME does not alter the loss or precondition updates, but filters stochastic gradients by projecting out a low-dimensional nuisance subspace associated with persistent high-variance directions. 

\section{Conclusion}\label{sec:conclusion}
We introduced \textbf{DOME}, an online first-order gradient filtering method that improves the signal-to-noise properties of stochastic optimization by removing structured, high-variance nuisance directions from stochastic gradients.
DOME tracks this nuisance subspace online using a streaming power method applied to the centered within-minibatch gradient covariance.
At each step, gradients are projected onto the orthogonal complement of the estimated subspace before being used in downstream operations, which reduces effective gradient norms without degrading learning dynamics in our experiments.
By explicitly separating informative descent directions from persistent high-variance fluctuations, DOME paves the way for more robust downstream uses of gradients, e.g. compression, continual learning (e.g., Elastic Weight Consolidation), and gradient-based analysis.

\section*{Impact Statement}
This paper presents work whose goal is to advance the field
of Machine Learning. There are many potential societal
consequences of our work, none which we feel must be
specifically highlighted here.

\clearpage

\bibliographystyle{icml2026}  
\bibliography{references}     
\newpage
\appendix
\onecolumn
\section{Overview}
The appendix contains background material used in the paper and additional experimental results.

\paragraph{Background.}
Appendix~\ref{app:la} recalls linear-algebra notions used in the algorithm description.

\paragraph{Additional experiments.}
Appendix~\ref{app:compression_sec} reports compression results on both MNIST and CIFAR-10.
Appendix~\ref{app:cov_spectrum_sec} presents the centered covariance spectra on MNIST and CIFAR-10.
Appendix~\ref{app:batchsize_sec} studies the effect of batch size on CIFAR-10, including accuracy trends and full learning dynamics.

\section{Linear algebra background}\label{app:la}
\paragraph{Eigenvalue Decomposition.} Let $\bm{A} \in \mathbb{R}^{n \times n}$ be a real-valued positive semi-definite matrix, where $n$ is a positive integer. The eigenvalue decomposition of $\bm{A}$ is given by $\bm{A} = \bm{U} \bm{\Lambda} \bm{U}^{\top}, $ where $\bm{U} \in \mathbb{R}^{n\times n}$ is a matrix of eigenvectors and $\bm{\Lambda}  \in \mathbb{R}^{n\times n}$ is a diagonal matrix containing corresponding eigenvalues.

\paragraph{QR Decomposition.}
We will use the matrix QR decomposition, obtained using the Gram-Schmidt procedure. Given a matrix $\bm{X} \in \mathbb{R}^{n \times p}$, the QR decomposition factorizes it as $\bm{X} = \bm{Q}\bm{R}$, where $\bm{Q} \in \mathbb{R}^{n\times p}$ is an orthonormal matrix (\emph{i.e.}, $\bm{Q}^{\top} \bm{Q} = \bm{I}$) and $\bm{R} \in \mathbb{R}^{p \times p}$ is an upper triangular matrix.

\paragraph{Random Gaussian Matrices.}
We denote by $\mathcal{N}(\mu, \sigma^2)^{n \times p}$ a $(n \times p)$ random matrix where each element is an independent and identically distributed (i.i.d.) random variable according to a Gaussian distribution with mean $\mu$ and variance $\sigma^2$.

\section{Training dynamics for a text classification task}
\label{app:text_classification}

We complement our image classification experiments with a text classification task, in order to assess whether the nuisance-subspace phenomenon identified by DOME also arises in a qualitatively different modality with sparse, high-dimensional inputs and a large number of classes.

\paragraph{Dataset.}
We use the \textbf{DBPedia-Classes} dataset, a hierarchical version of the DBPedia ontology classification benchmark.\footnote{\url{https://huggingface.co/datasets/DeveloperOats/DBPedia_Classes}}
Each example consists of a short textual description of a Wikipedia entity, associated with labels at three levels of semantic granularity. We focus on the \textbf{L2 level}, which contains \textbf{70 classes} and exhibits a non-uniform class distribution. We use the standard train/validation/test splits provided with the dataset.

\paragraph{Model and optimization.}
We train a lightweight Transformer encoder from scratch, without any pretrained embeddings.
The model consists of a word-level embedding layer followed by three Transformer encoder layers with hidden dimension $d=64$, $4$ attention heads, and a feedforward dimension of $256$, similar to the architecture used in \citet{damian2023self}.
A linear classifier is applied to the \texttt{[CLS]} token representation.
Training is performed for $5$ epochs using SGD with learning rate $10^{-4}$ and batch size $512$, and $k=C$.
All experiments are repeated over $5$ random seeds.

\paragraph{Evaluation metric.}
Due to the class imbalance at the L2 level, we report \textbf{macro-F1} rather than accuracy.
Macro-F1 averages the per-class F1 scores and therefore weights all classes equally, preventing dominant classes from masking failures on underrepresented categories. This choice is standard for large-scale multi-class text classification tasks with skewed label distributions.

\begin{figure}[H]
  \centering
  \begin{subfigure}[t]{0.33\linewidth}
    \centering
    \includegraphics[width=\linewidth]{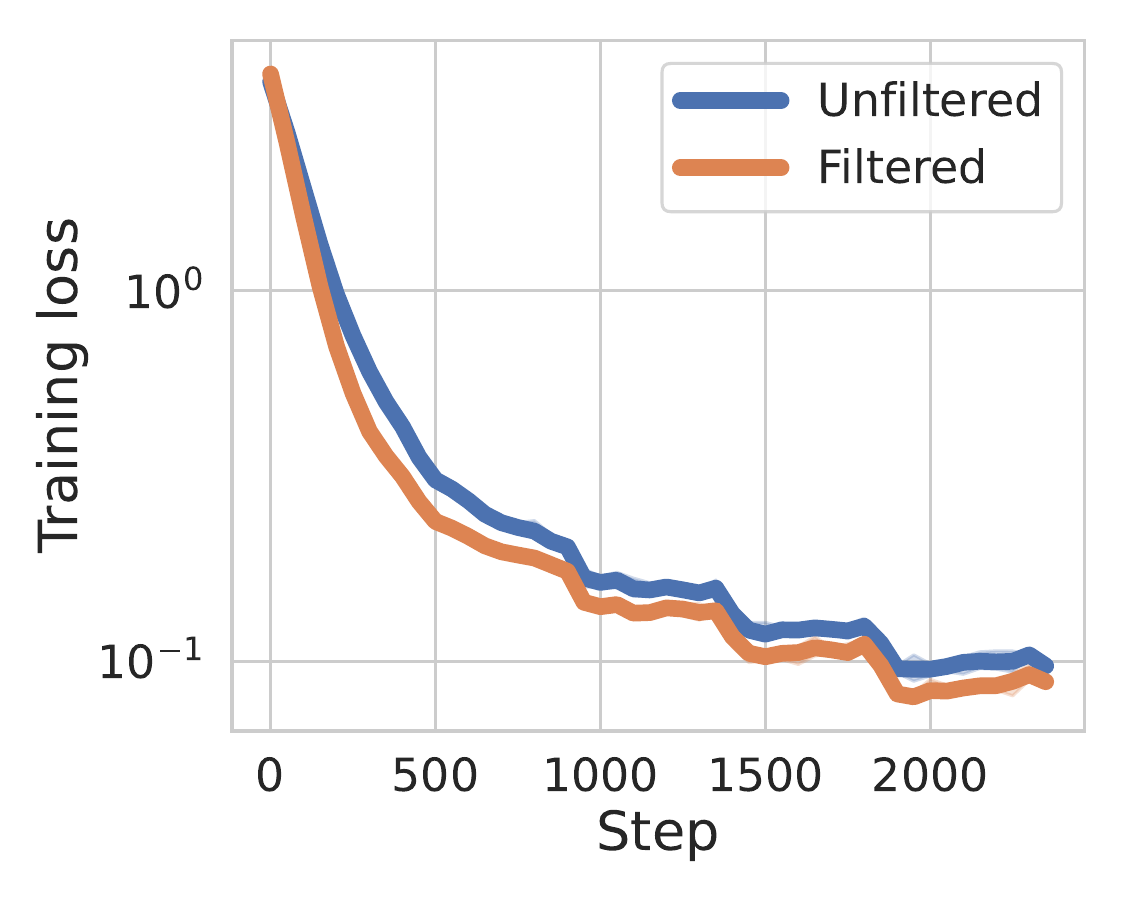}
    \label{fig:training_dbpedia_loss}
  \end{subfigure}\hfill
  \begin{subfigure}[t]{0.33\linewidth}
    \centering
    \includegraphics[width=\linewidth]{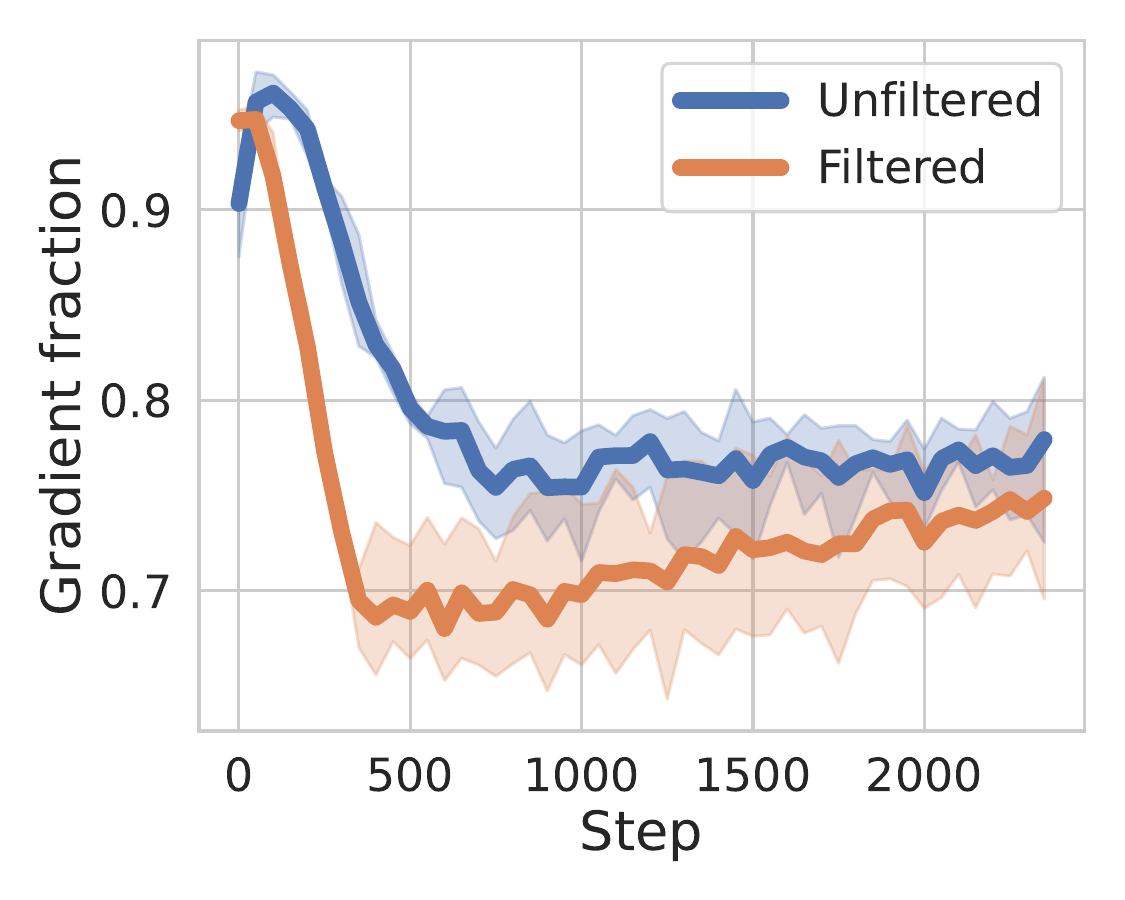}
    \label{fig:training_dbpedia_fraction}
  \end{subfigure}\hfill
  \begin{subfigure}[t]{0.33\linewidth}
    \centering
    \includegraphics[width=\linewidth]{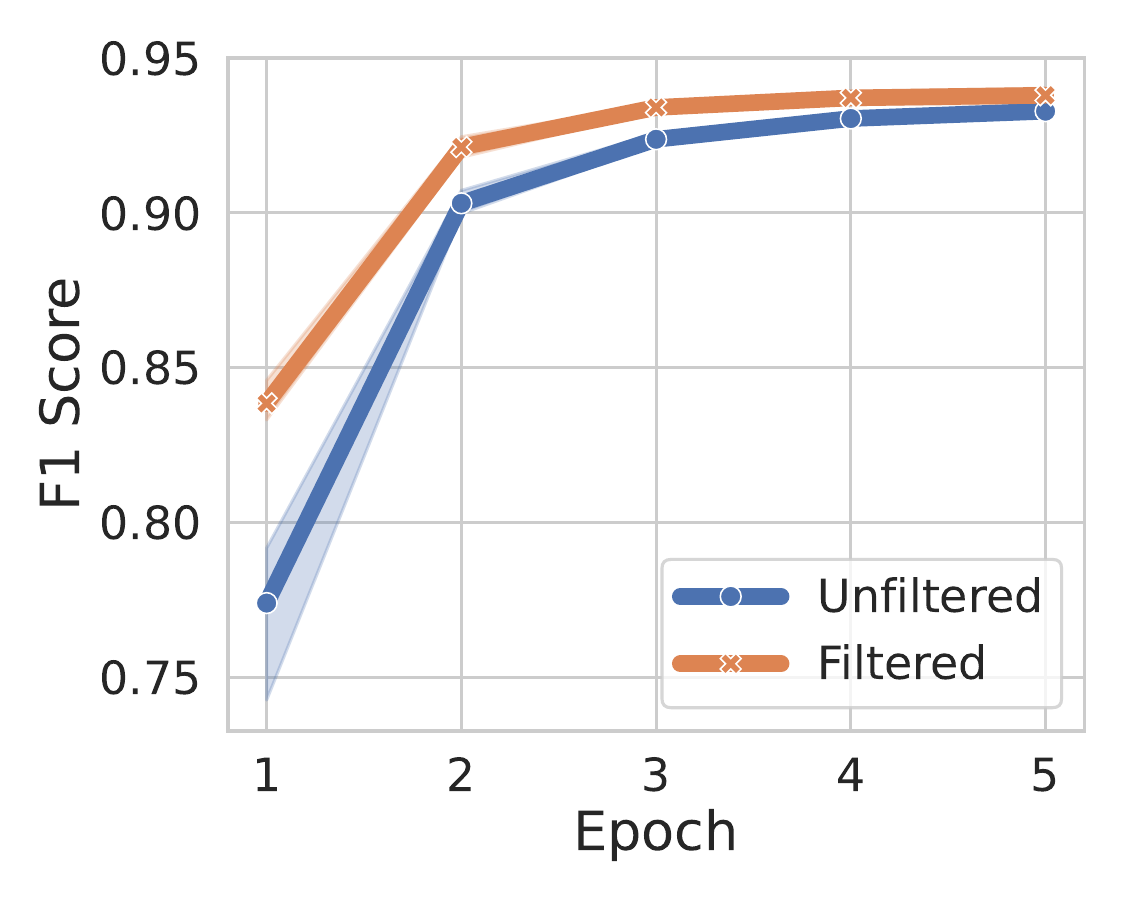}
    \label{fig:training_dbpedia_f1}
  \end{subfigure}
  \vspace{-4mm}
  \caption{
  \textbf{Impact of filtering on training dynamics on DBPedia-Classes-L2} when training a 3-layer Transformer with SGD, $lr=10^{-4}$ and a batch size of 512 for 5 epochs.
  \textbf{Left}: Training loss as a function of epochs for the unfiltered optimizer and its filtered counterpart.
  \textbf{Center}: Average fraction of gradient norms lying in the dominant subspace (before applying filtering for the filtered version).
  \textbf{Right}: F1-score as a function of epochs.
  Shaded areas indicate $95\%$ bootstrap confidence intervals over 5 random seeds.
  }
  \label{fig:training_dpbedia}
    \vspace{-2mm}
\end{figure}

\paragraph{Results.}
Figure~\ref{fig:training_dpbedia} reports the training loss, the fraction of gradient norm captured by the dominant covariance subspace, and the macro-F1 score over training epochs, comparing standard SGD with its DOME-filtered counterpart.

As in the image classification setting, we observe that a substantial fraction of the gradient norm concentrates in a low-dimensional subspace estimated from the centered within-minibatch gradient covariance.
Removing this dominant subspace at each iteration also does not degrade optimization performance.
The filtered counterpart achieves slightly higher macro-F1 score in the first epochs but performance for both approaches becomes similar as training progresses.

These results demonstrate that the presence of this dominant, high-variance gradient subspace is not specific to convolutional architectures or image data.
Even in a text classification task with sparse inputs, many classes, and a fundamentally different inductive bias, removing the leading covariance directions has little effect on convergence or generalization.
This further supports our interpretation of these directions as a nuisance subspace capturing persistent stochastic fluctuations rather than essential descent information.
\section{Impact of compression} \label{app:compression_sec}
\begin{figure}[H]
  \centering
  \begin{subfigure}[t]{0.49\linewidth}
    \centering
    \includegraphics[width=\linewidth]{results/mnist_compression.pdf}
    \caption{MNIST}
    \label{app:compression_mnist}
  \end{subfigure}\hfill
  \begin{subfigure}[t]{0.49\linewidth}
    \centering
    \includegraphics[width=\linewidth]{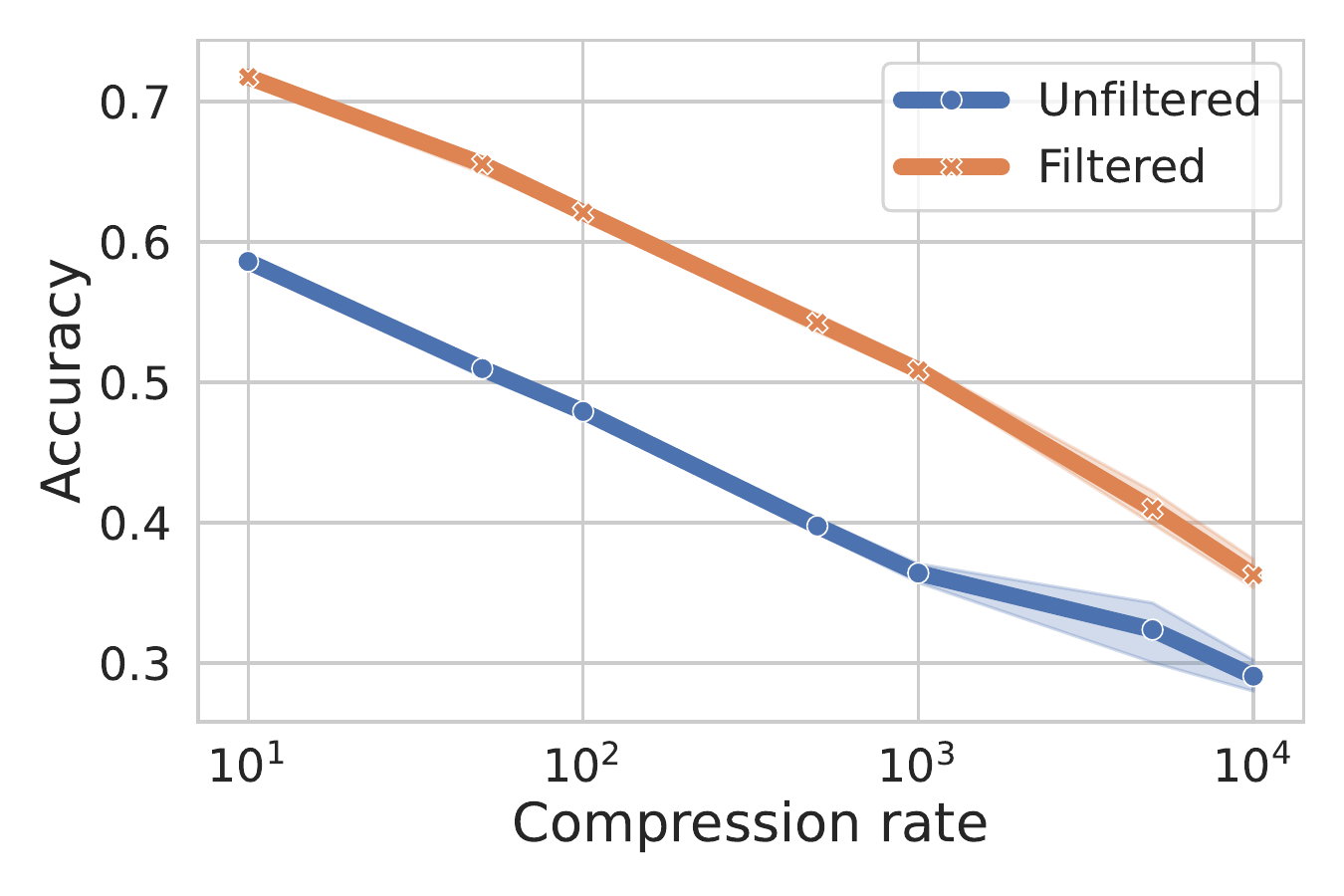}
    \caption{CIFAR-10}
    \label{app:compression_cifar}
  \end{subfigure}
  \caption{\textbf{Compression tolerance.} Test accuracy versus compression rate for standard Adam and DOME-filtered Adam using random Gaussian projection sketches. Curves report means over 5 runs; shaded regions indicate 95\%
  bootstrap confidence intervals.}
  \label{app:compression}
\end{figure}

Figure \ref{app:compression} reports additional compression results on MNIST and CIFAR-10, complementing the main text by showing the accuracy–compression trade-off for both unfiltered and DOME-filtered Adam.

\section{Covariance Spectrum}\label{app:cov_spectrum_sec}
\begin{figure}[H]
  \centering
  \begin{subfigure}[t]{0.5\linewidth}
    \centering
    \includegraphics[width=\linewidth]{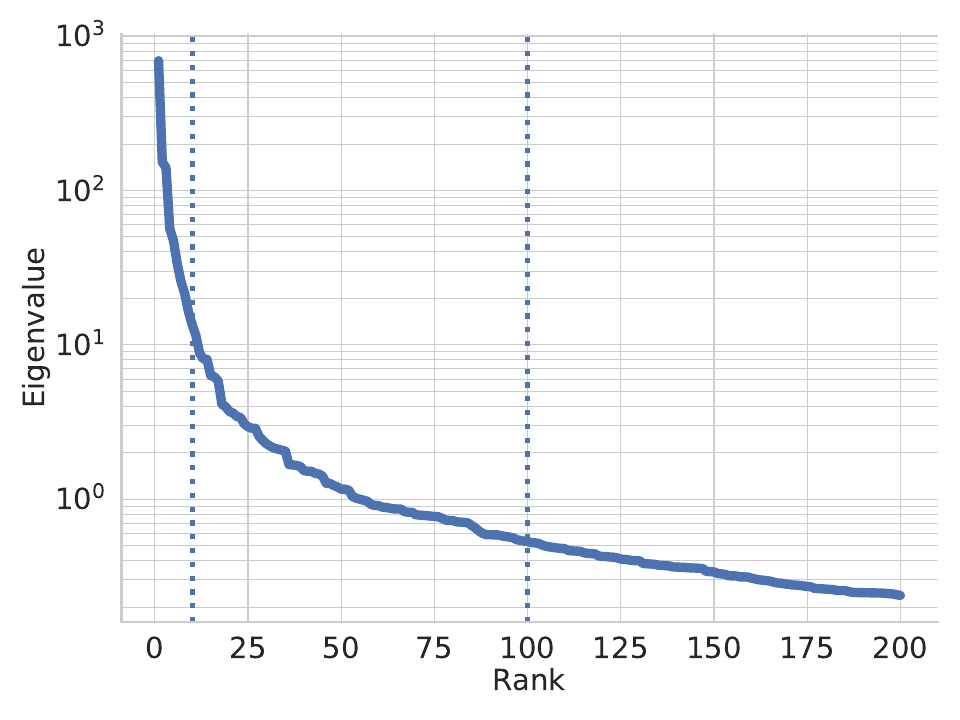}
    \caption{MNIST}
    \label{app:mnist_spectrum}
  \end{subfigure}
  \begin{subfigure}[t]{0.49\linewidth}
    \centering
    \includegraphics[width=\linewidth]{results/cifar10_spectrum.pdf}
    \caption{CIFAR-10}
    \label{app:cifar10_spectrum}
  \end{subfigure}
  \caption{\textbf{Centered covariance spectrum.} Eigenvalue spectrum of the centered gradient covariance after training for 10 epochs on MNIST
  (left) and CIFAR-10 (right). The first dotted line is at rank 10 and the
  second at rank 100.}
  \label{app:cov_spectrum}
\end{figure}

Figure \ref{app:cov_spectrum} shows the eigenvalue spectrum of the centered gradient covariance on MNIST and CIFAR-10, highlighting the presence of a small set of dominant outliers followed by a broad bulk.

\section{Impact of the batch size}\label{app:batchsize_sec}
\begin{figure}[H]
\centering
\includegraphics[width=0.5\linewidth]{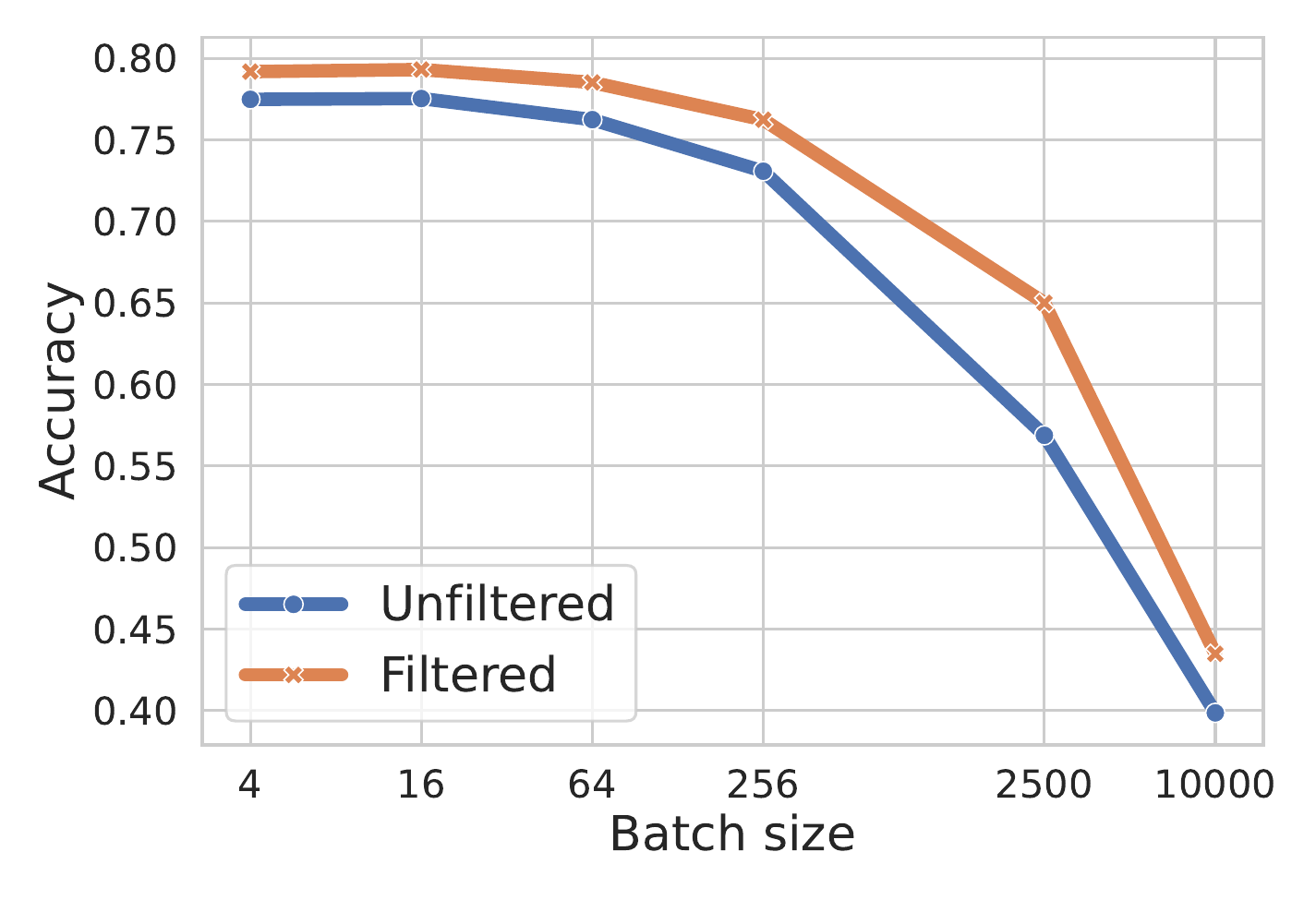}
\caption{\textbf{CIFAR-10 accuracy vs. batch size.}
Test accuracy as a function of the batch size (log scale), comparing
\textit{DOME-Filtered} training, and \textit{Unfiltered} training. Each point corresponds to the best epoch selected after averaging accuracy across random seeds for
each epoch. Shaded regions indicate 99\% bootstrap confidence intervals.}
\label{app:cifar10_acc_vs_batchsize}
\end{figure}

\begin{figure*}[t]
  \centering

  \noindent
  \begin{minipage}[c]{0.03\linewidth}
    \centering
    \scriptsize\rotatebox{90}{\textbf{Batch size: 4}}
  \end{minipage}\hfill
  \begin{minipage}[c]{0.96\linewidth}
    \includegraphics[width=0.31\linewidth]{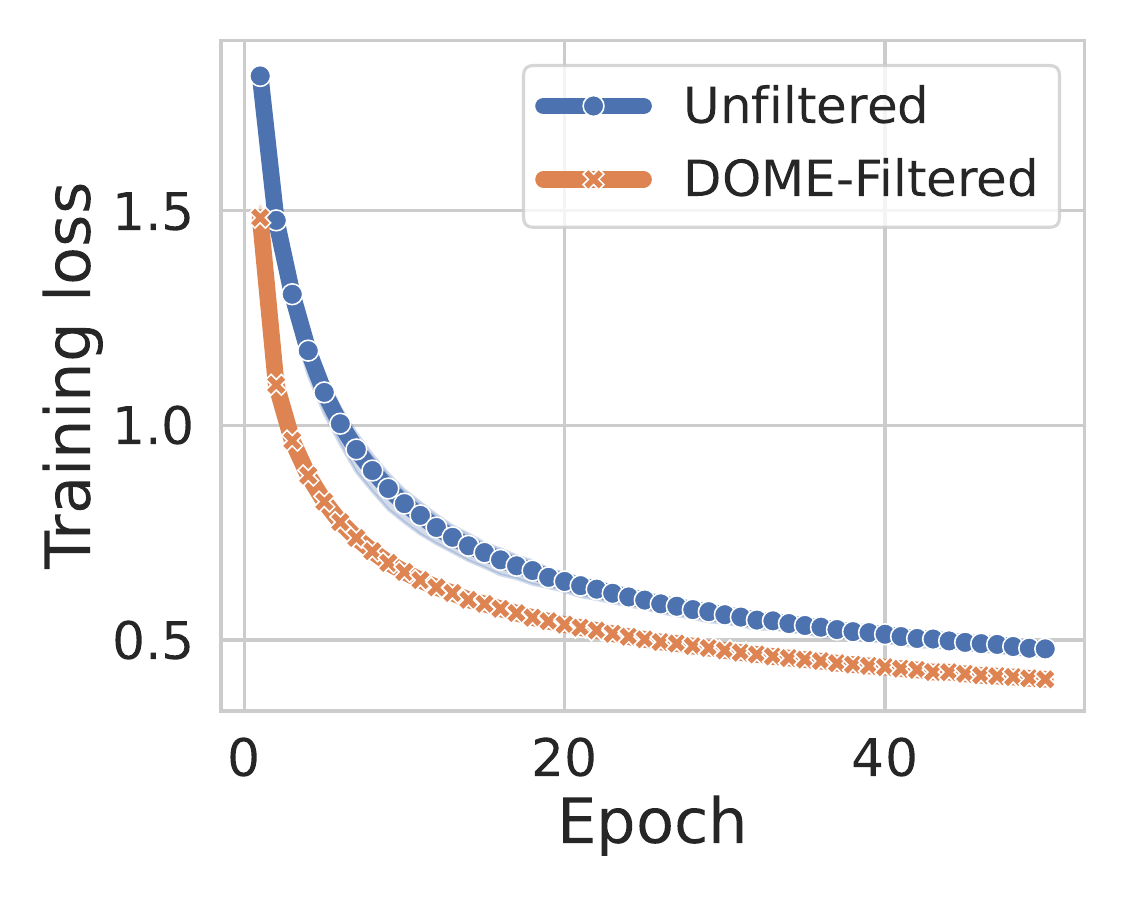}\hfill
    \includegraphics[width=0.31\linewidth]{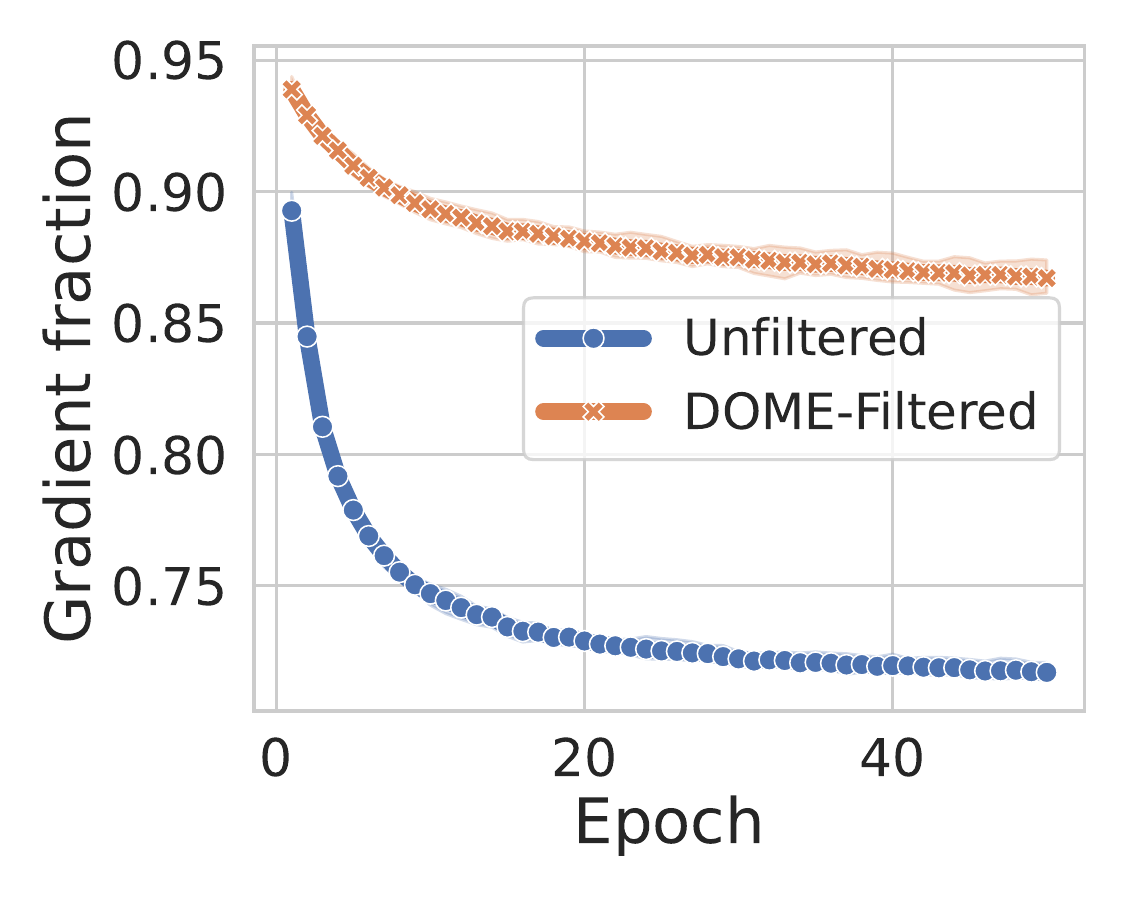}\hfill
    \includegraphics[width=0.31\linewidth]{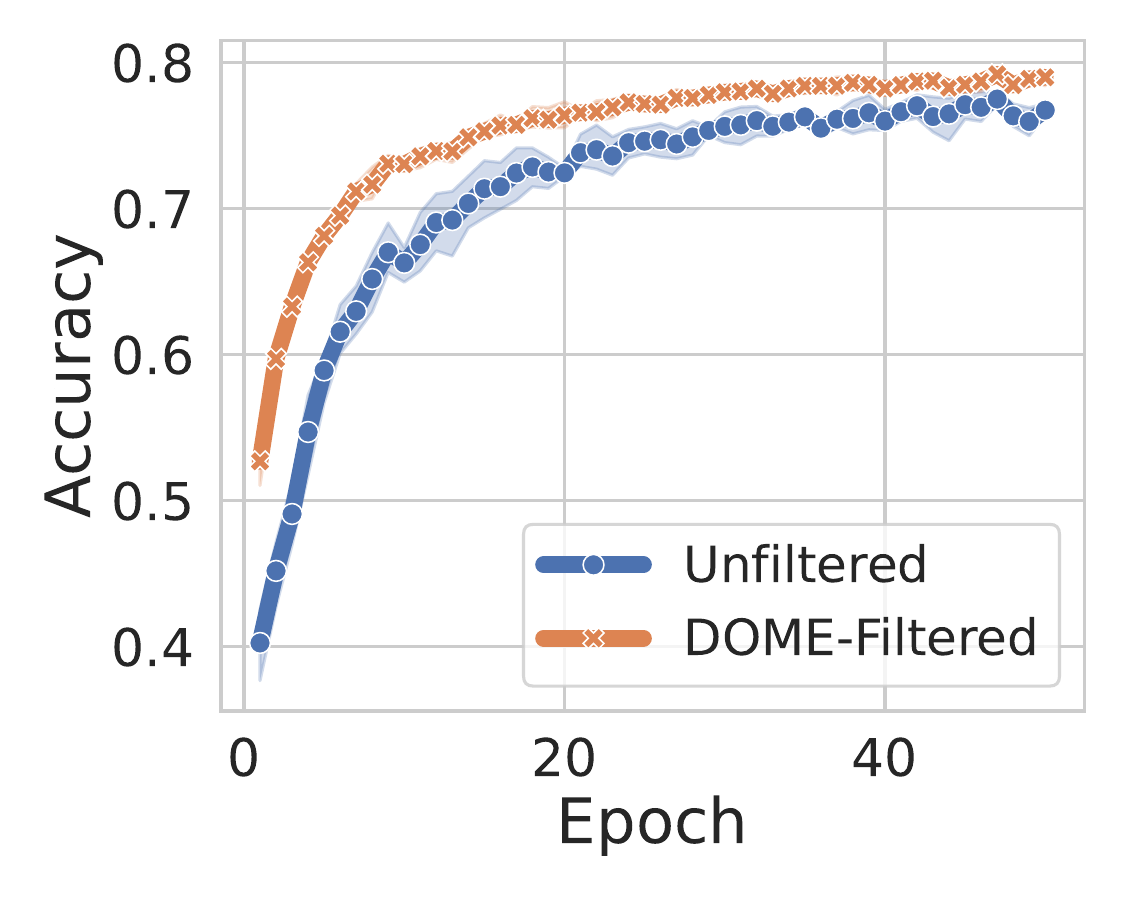}
  \end{minipage}

  \vspace{1.2mm}

  \noindent
  \begin{minipage}[c]{0.03\linewidth}
    \centering
    \scriptsize\rotatebox{90}{\textbf{Batch size: 64}}
  \end{minipage}\hfill
  \begin{minipage}[c]{0.96\linewidth}
    \includegraphics[width=0.31\linewidth]{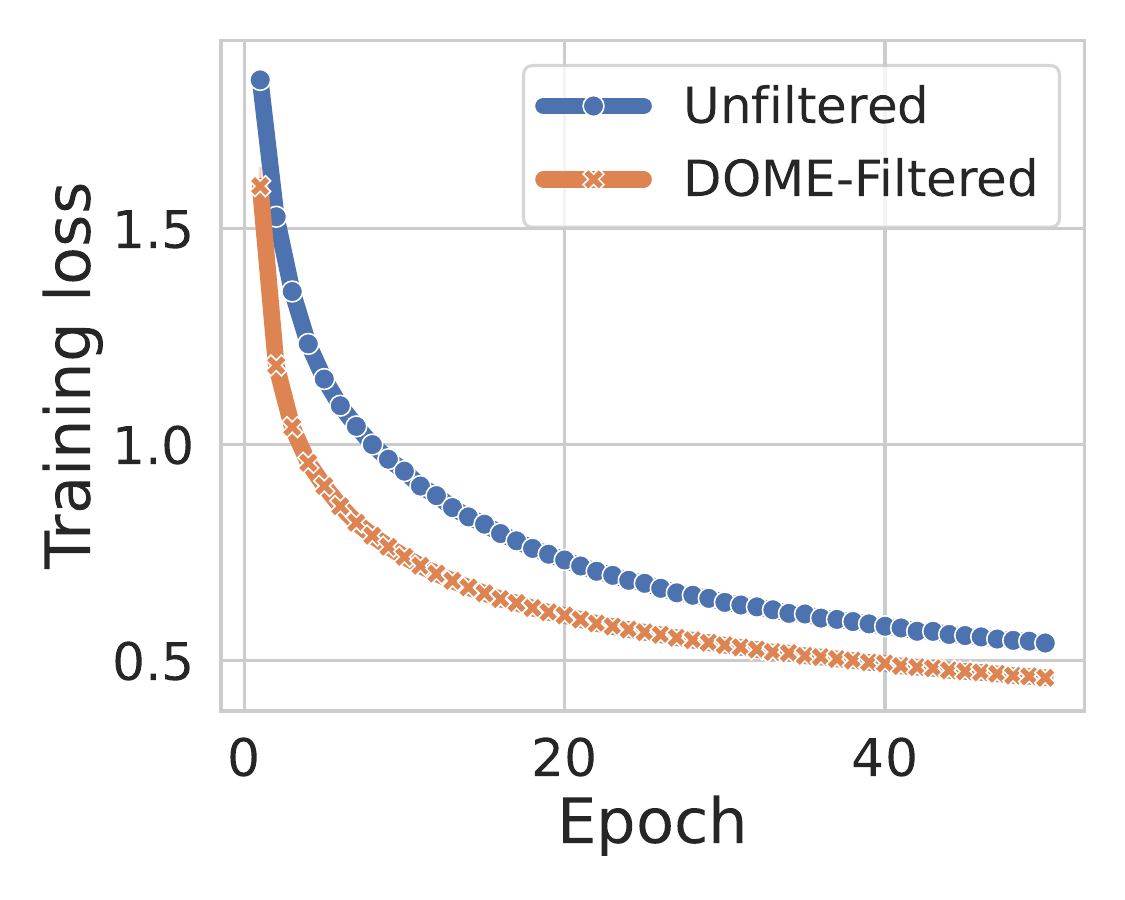}\hfill
    \includegraphics[width=0.31\linewidth]{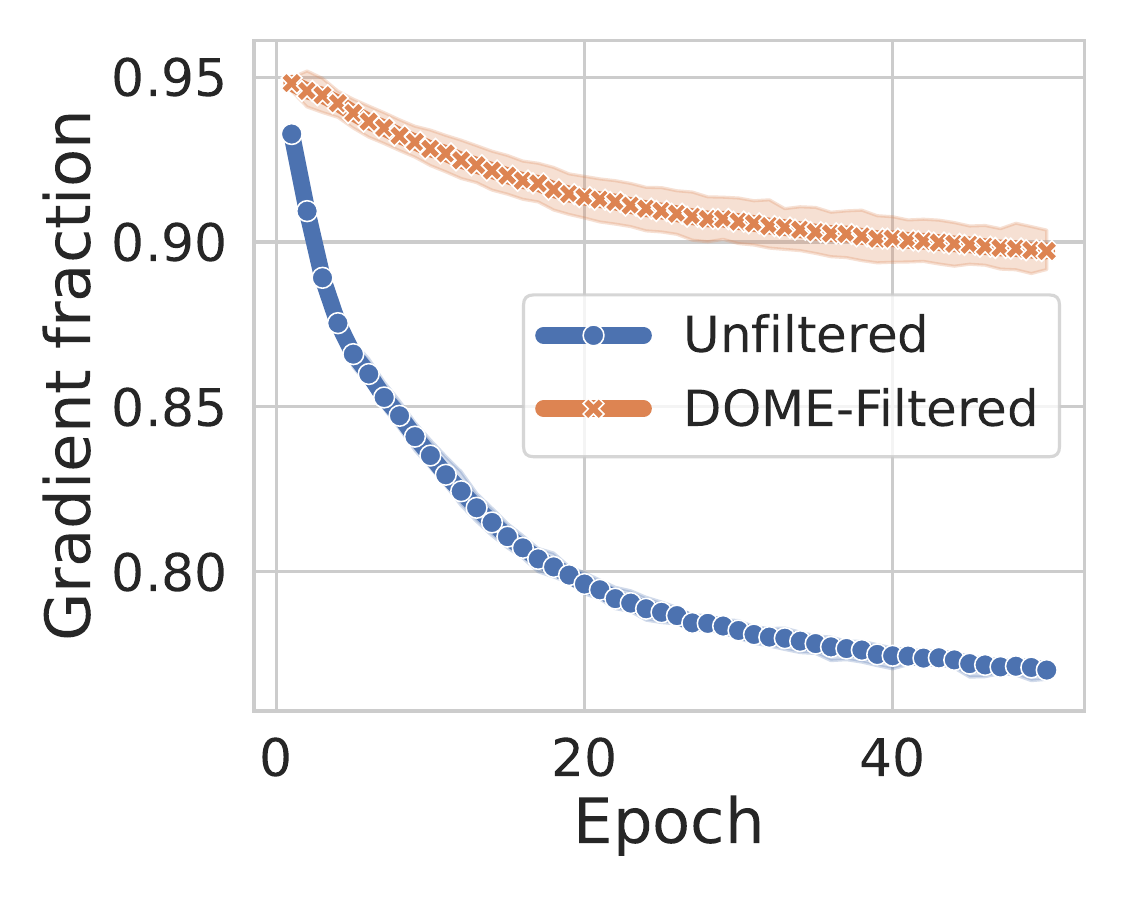}\hfill
    \includegraphics[width=0.31\linewidth]{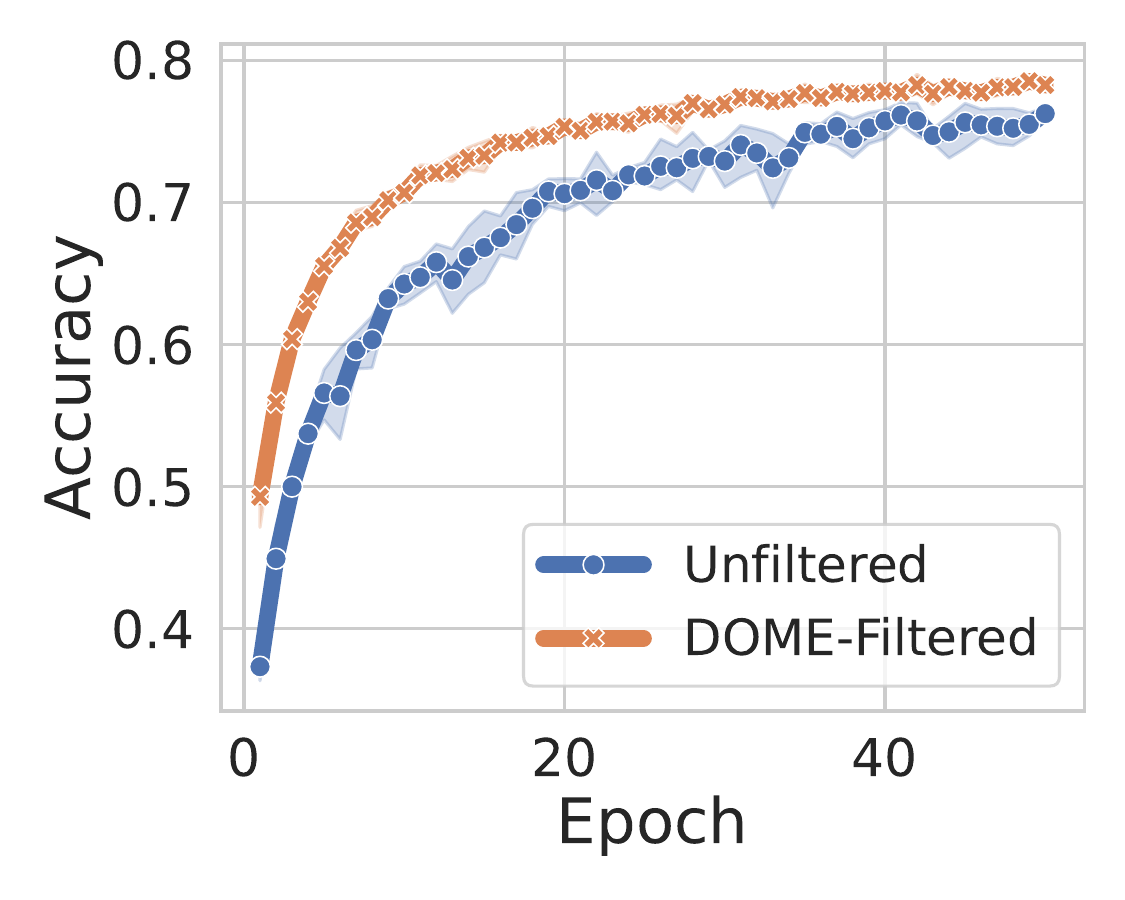}
  \end{minipage}

  \vspace{1.2mm}

  \noindent
  \begin{minipage}[c]{0.03\linewidth}
    \centering
    \scriptsize\rotatebox{90}{\textbf{Batch size: 256}}
  \end{minipage}\hfill
  \begin{minipage}[c]{0.96\linewidth}
    \includegraphics[width=0.31\linewidth]{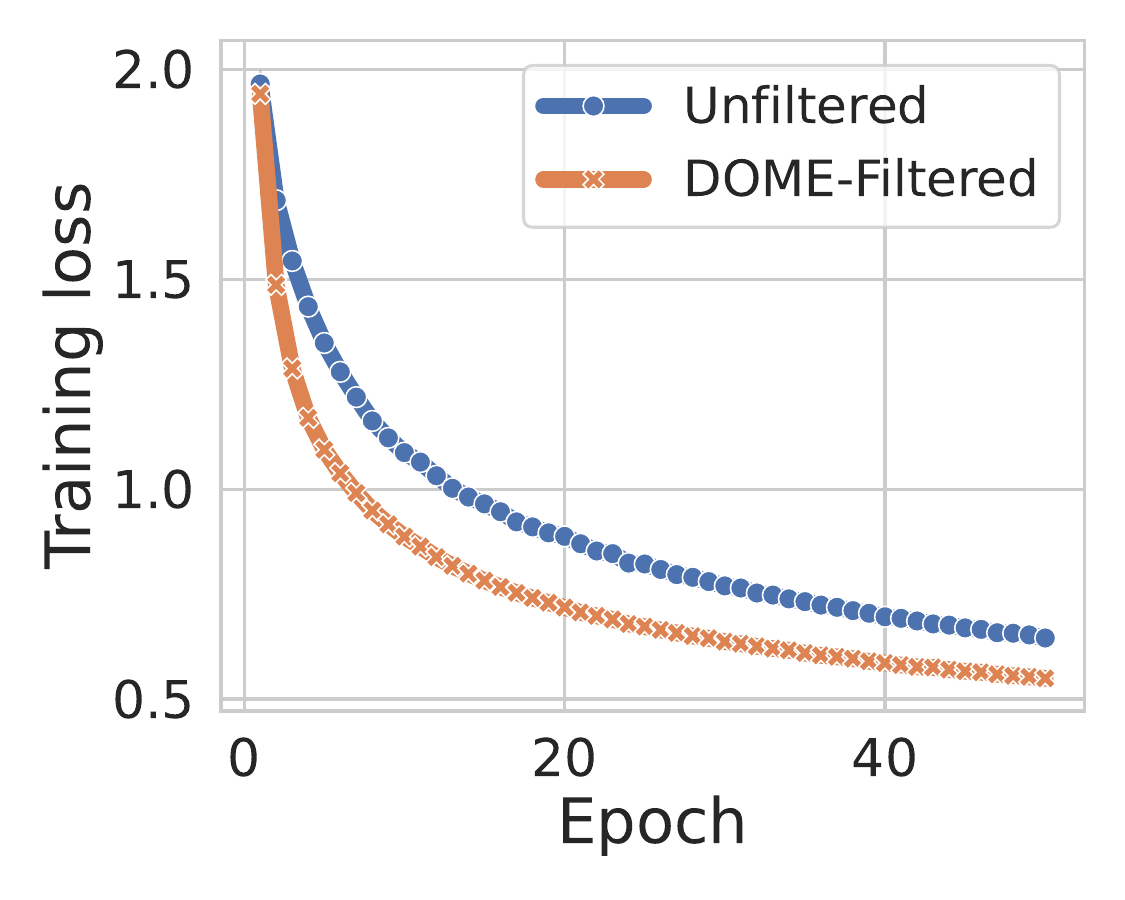}\hfill
    \includegraphics[width=0.31\linewidth]{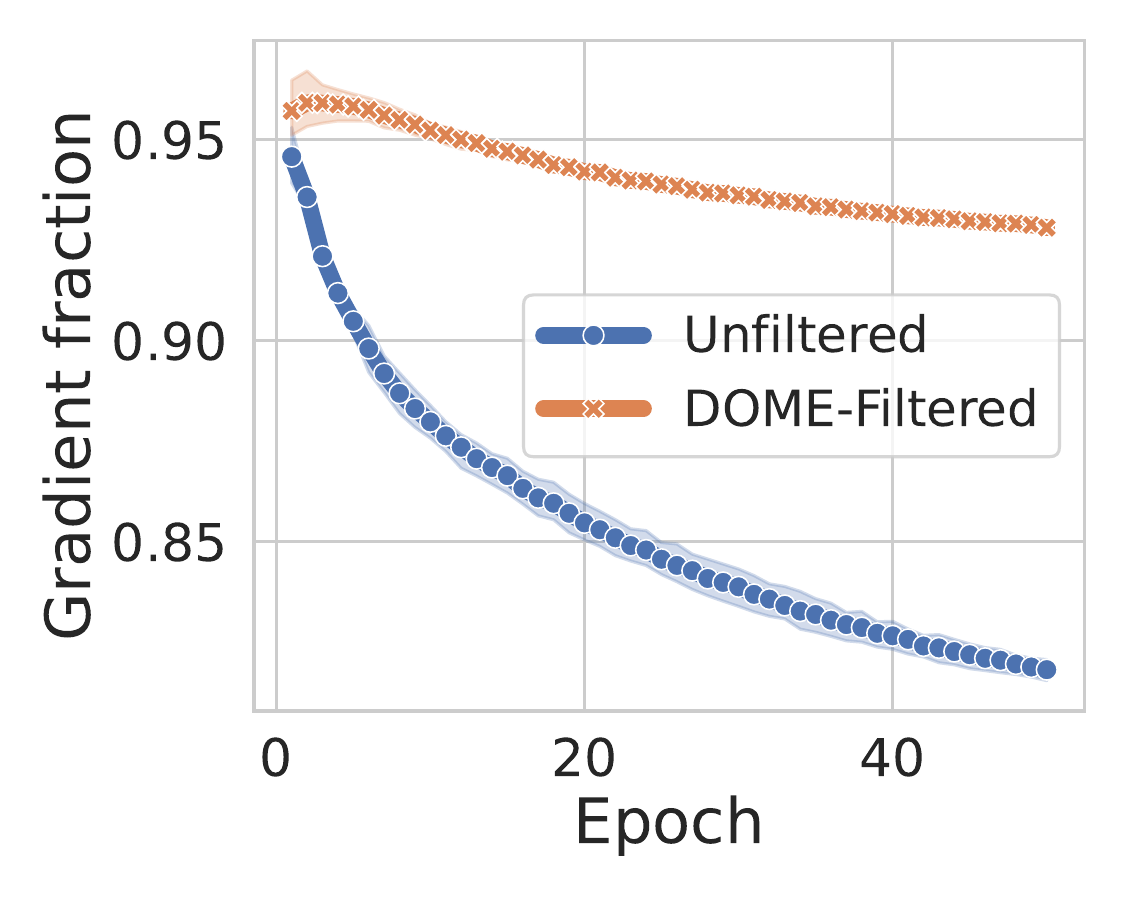}\hfill
    \includegraphics[width=0.31\linewidth]{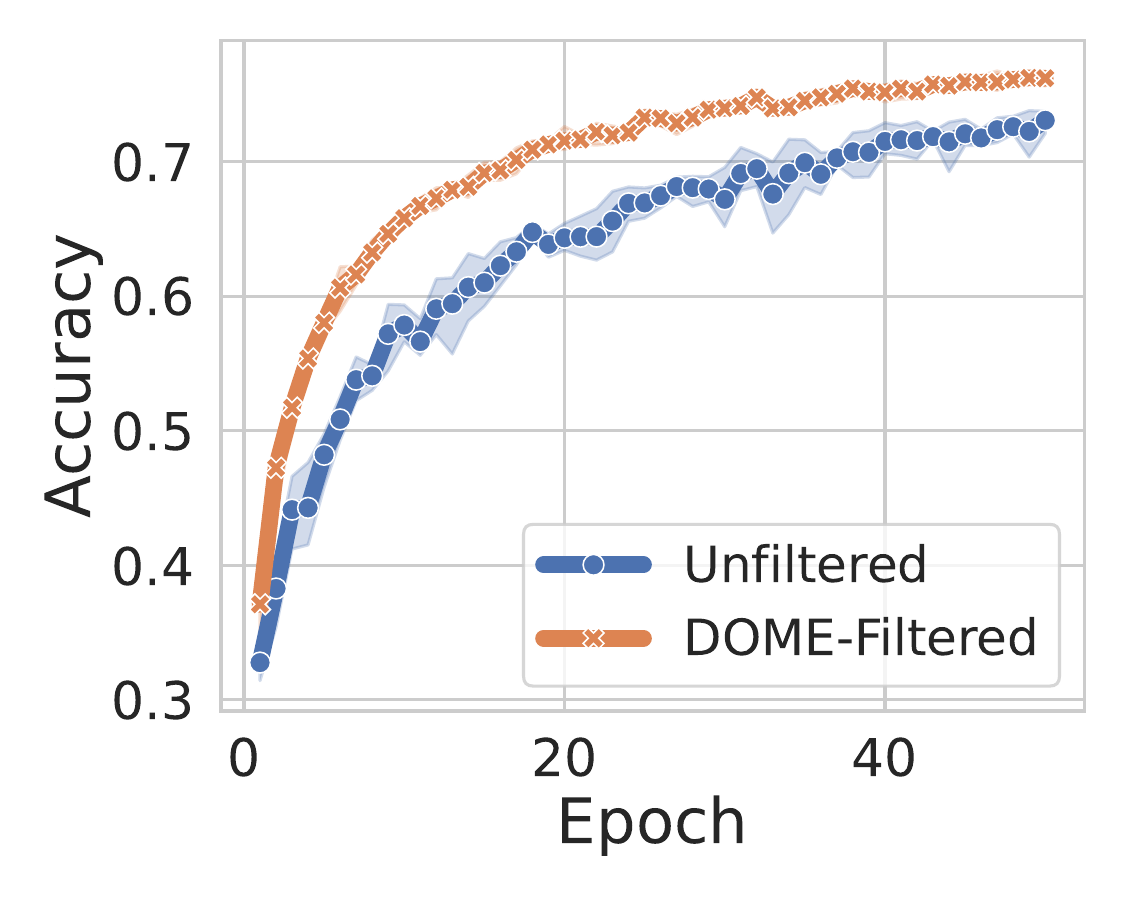}
  \end{minipage}

  \vspace{1.2mm}

  \noindent
  \begin{minipage}[c]{0.03\linewidth}
    \centering
    \scriptsize\rotatebox{90}{\textbf{Batch size: 2500}}
  \end{minipage}\hfill
  \begin{minipage}[c]{0.96\linewidth}
    \includegraphics[width=0.31\linewidth]{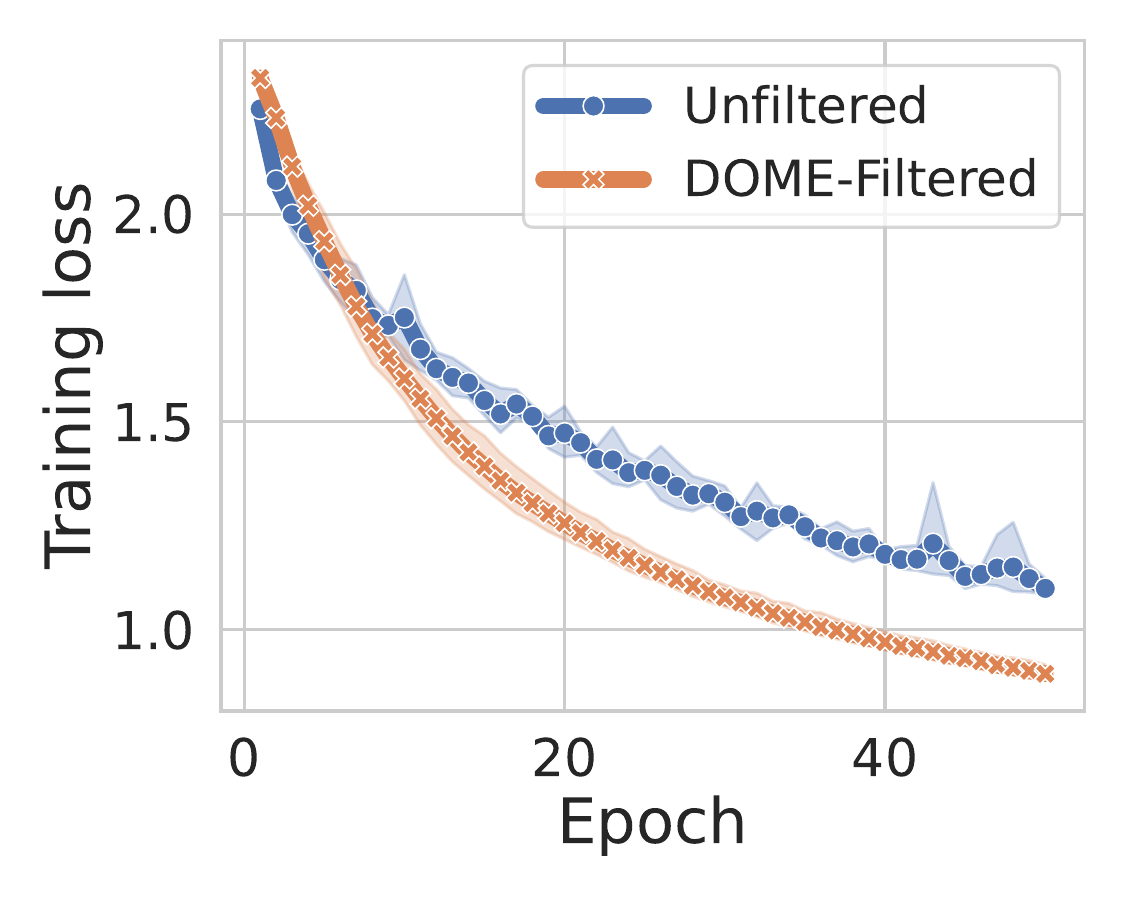}\hfill
    \includegraphics[width=0.31\linewidth]{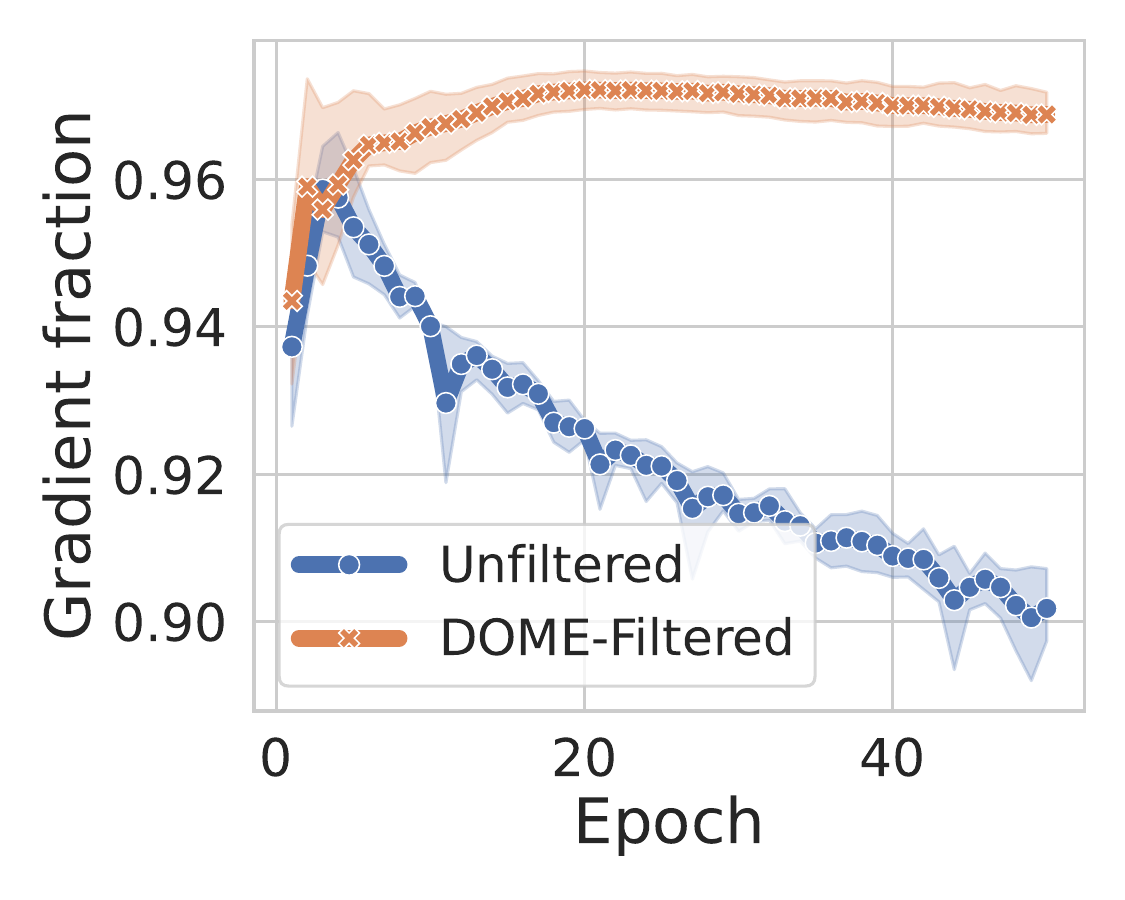}\hfill
    \includegraphics[width=0.31\linewidth]{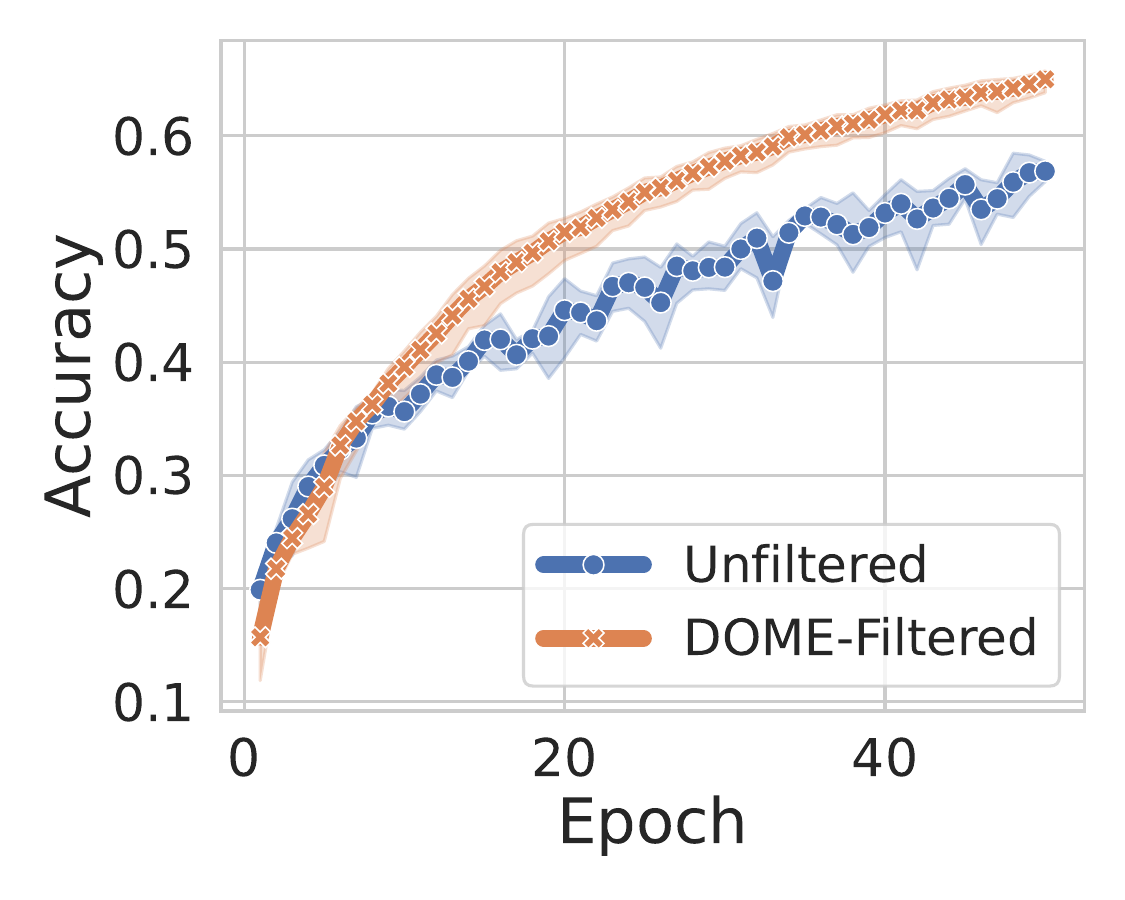}
  \end{minipage}

  \vspace{1.2mm}

  \noindent
  \begin{minipage}[c]{0.03\linewidth}
    \centering
    \scriptsize\rotatebox{90}{\textbf{Batch size: 10000}}
  \end{minipage}\hfill
  \begin{minipage}[c]{0.96\linewidth}
    \includegraphics[width=0.31\linewidth]{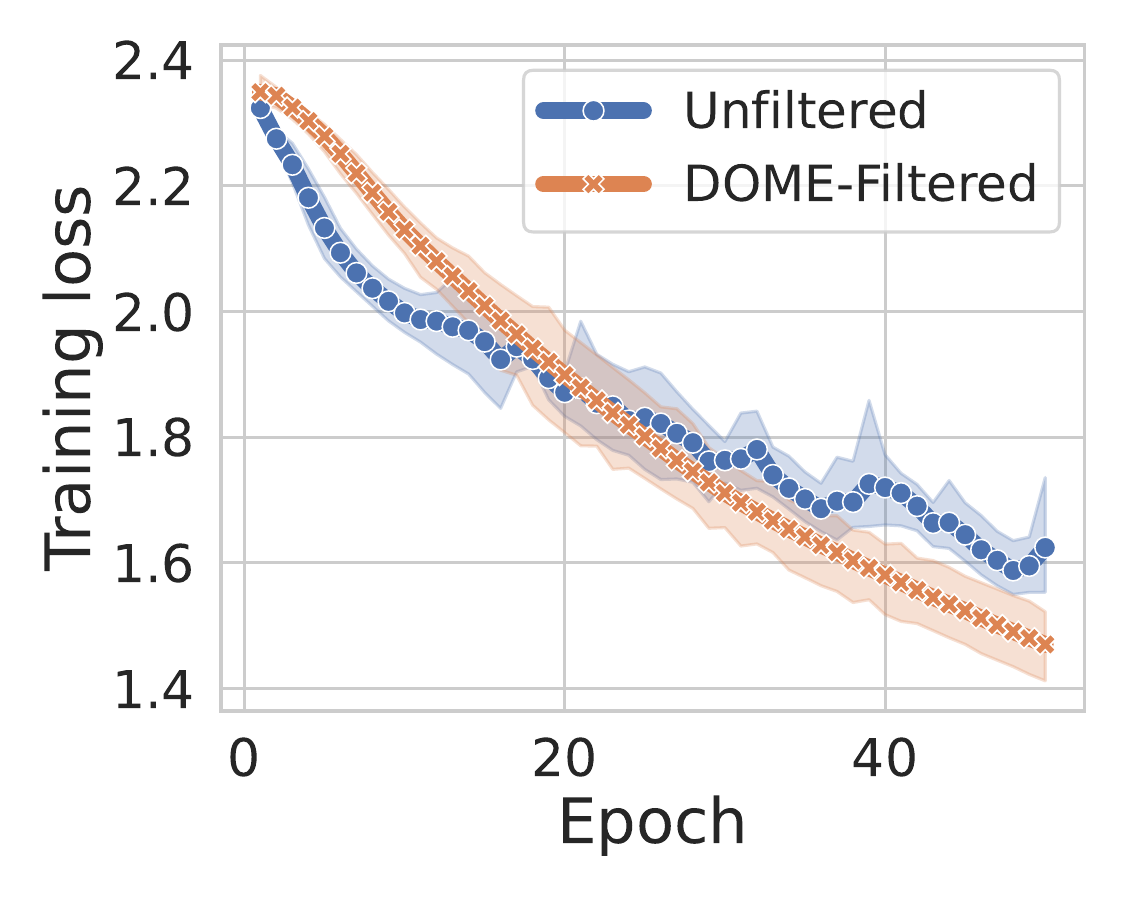}\hfill
    \includegraphics[width=0.31\linewidth]{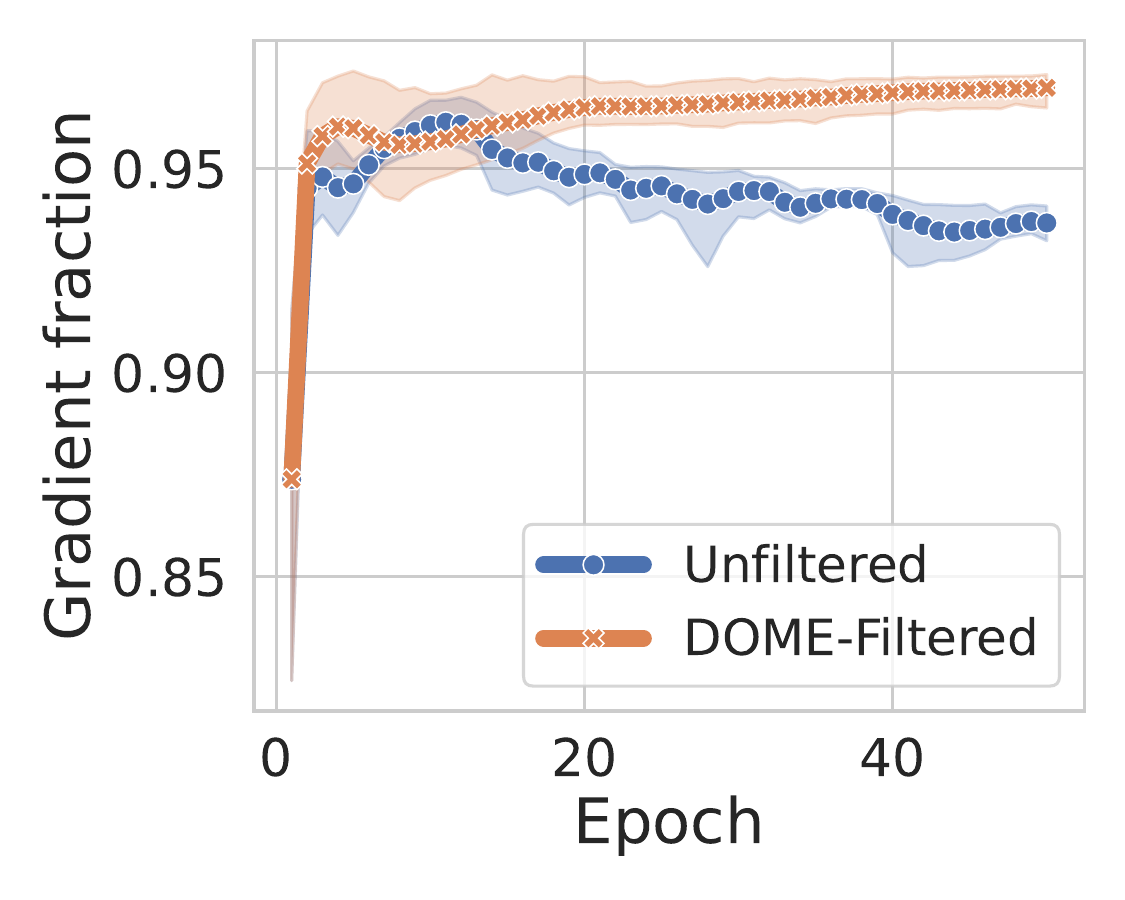}\hfill
    \includegraphics[width=0.31\linewidth]{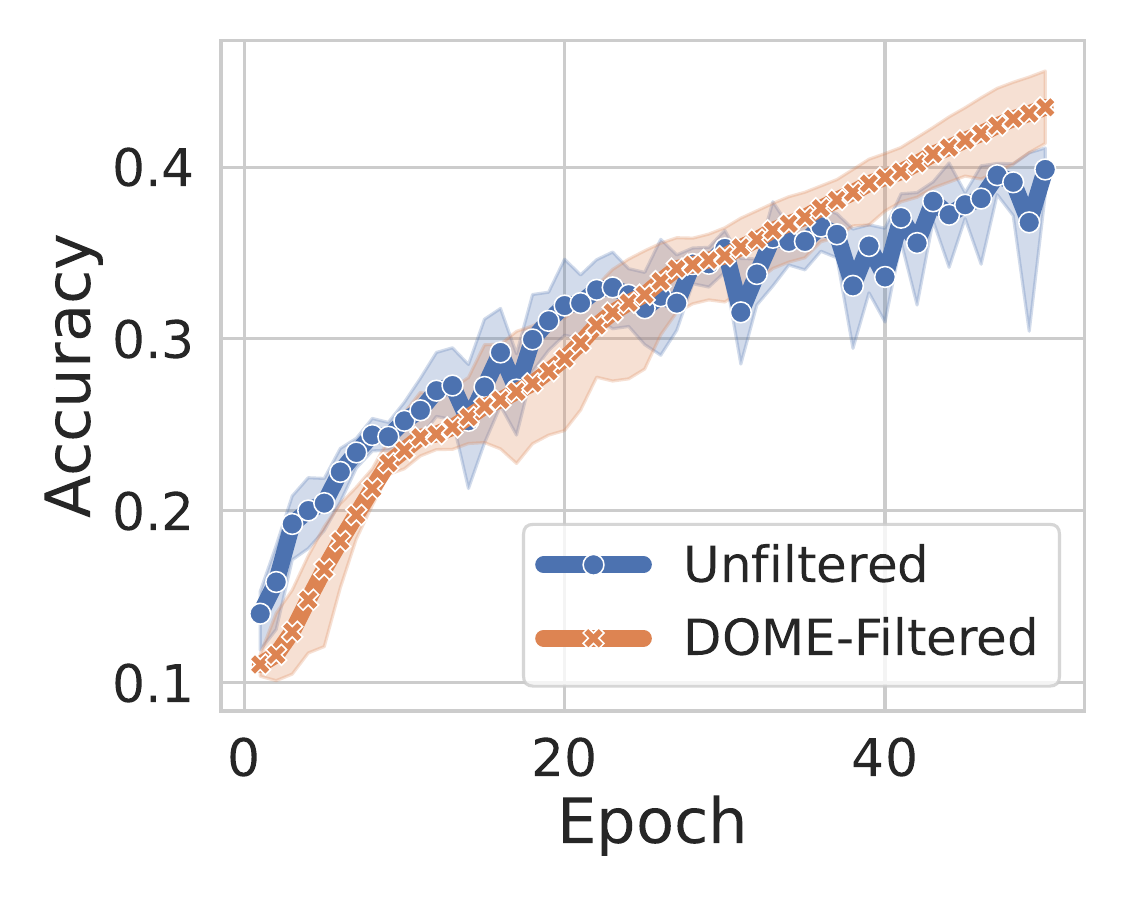}
  \end{minipage}

  \vspace{-2.5mm}
  \caption{
  \textbf{Impact of DOME filtering on CIFAR-10 training dynamics across batch sizes.}
  Each row corresponds to a different batch size.
  \textbf{Left}: training loss vs.\ epoch.
  \textbf{Center}: fraction of gradient norm captured by the dominant subspace.
  \textbf{Right}: top-1 accuracy vs.\ epoch.
  Shaded regions indicate 99\% bootstrap confidence intervals over random seeds.
  }
  \label{app:cifar10_training_multibs}
\end{figure*}

Figure \ref{app:cifar10_acc_vs_batchsize} reports CIFAR-10 test accuracy as a function of batch size, comparing DOME-filtered and unfiltered training across a wide range of regimes.
Figure \ref{app:cifar10_training_multibs} provides the full training dynamics on CIFAR-10 for several batch sizes, including loss, dominant-subspace fraction, and accuracy trajectories. We see that our analysis from Experiment~\ref{learndynamics} extends to other batch sizes.

\end{document}